\documentclass[aps,pre,onecolumngroupedaddress]{revtex4-2}

\usepackage{amsmath,amssymb}
\usepackage{newtxtext,newtxmath}
\usepackage{algorithm}
\usepackage{algpseudocode}
\usepackage{bm,url}

\usepackage{graphicx,graphics}

\def\newblock{\hskip .11em plus .33em minus .07em}

\algnewcommand{\Initialize}[1]{%
  \State \textbf{Initialize:}
  \State \hspace*{\algorithmicindent}\parbox[t]{0.8\linewidth}{\raggedright #1}
}

\newcommand{\Fref}[1]{Fig.\ref{#1}}
\newcommand{\eref}[1]{eq.(\ref{#1})}

\begin{document}

\title{Bayesian Inference of Infected Patients in Group Testing\\
with Prevalence Estimation}

\author{Ayaka Sakata}
\email[]{ayaka@ism.ac.jp}
\affiliation{Institute of Statistical Mathematics, 
10-3 Midori-cho, Tachikawa, Tokyo 190-8562, Japan}
\affiliation{
Department of Statistical Science, 
The Graduate University for Advanced Science
(SOKENDAI), Hayama-cho, Kanagawa 240-0193, Japan}
\affiliation{
JST PRESTO, 4-1-8 Honcho, Kawaguchi, Saitama, 332-0012, Japan} 

\begin{abstract}
Group testing is a method of identifying 
infected patients by performing tests on a pool of specimens collected from patients.
For the case in which the test returns a false result with
finite probability, Bayesian inference 
and a corresponding belief propagation (BP) algorithm are introduced
to identify the infected patients from the results of tests performed on the pool.
It is shown that the true-positive rate is improved by taking into account 
the credible interval of a point estimate of each patient.
Further, the prevalence and the error probability in the test 
are estimated by combining an expectation-maximization method with the BP algorithm.
As another approach, a hierarchical Bayes model is introduced
to identify the infected patients and estimate the prevalence.
By comparing these methods, a guide for practical usage is formulated.
\end{abstract}

\maketitle

\section{Introduction}

In clinical testing methods such as blood tests and polymerase chain reaction (PCR) tests,
discovering infected patients
from a large population requires significant operating costs.
Because of limitations in
the number of available devices, reagents, and technologists,
a high demand exists for more efficient methods of testing.
Group testing is one of the approaches 
for the reduction of the operating costs
by performing tests on pools of specimens
obtained from patients \cite{Dorfman,GT_book}.
It is known that if the rate of the infected patients 
in the population is sufficiently small, in principle,
one can identify the infected patients
from the tests on pools whose number is smaller than that of the population.
Originally, group testing was developed for 
blood testing during World War II,
and is now applied to various fields 
such as quality control in product testing \cite{product},
estimation of the content of 
genetically mutated organisms in maize grains \cite{GM_GT},
and multiple access communication \cite{Wolf}.

The identification of the infected patients from the 
results of group test is 
mathematically formulated as a channel coding problem
to reconstruct the original signal from a codeword
transferred through noisy channel \cite{Katona},
where the original signal, codeword, and noisy channel correspond to 
the state of patients, states of the pools, and errors in the tests,
respectively.
Further, group testing can be regarded as a variant of compressed sensing \cite{Candes-Tao,Donoho2006}
with discrete variables and logical sums.
Hence, the progress in the last decade in sparse estimation 
including compressed sensing has
revived interest in group testing,
and information-theory approaches have achieved bound evaluation of group testing at a limiting case \cite{Atia,Johnson}.

Recently, in response to the epidemic infection of COVID-19 
that requires testing on large populations,
the idea of group testing has attracted increasing attention \cite{COVID-GT1,COVID-GT2,COVID-GT3}
from the viewpoint of practical application rather than mathematics.
In practice, clinical testing sometimes results in errors
even when the operation is precise.
For example, in PCR tests, false negative probabilities of
up to 3\% and false positive (FP) probabilities of up to 5\% have been observed \cite{BayesianGT}.
Moreover, the bacterial or viral load
in the specimen
depends on the specimen collection method and timing
\cite{COVID19}.
Therefore, a specimen sometimes does not contain
a sufficient amount of the pathogen
to exceed the detection limit,
even when the patient is infected.
Further, post-collection contamination of pathogens into a specimen
can cause a positive result even when the patient is not infected.
%
Statistical inference can contribute to
the correction of errors by estimating the true state of patients from
noisy test data, and 
quantifying the credibility of the estimation.

In this paper, Bayesian inference is introduced
to identify the infected patients in the
group testing problem
considering the finite false probabilities in the test.
The infection probability of each patient is 
approximately calculated by a belief propagation (BP) algorithm
because of its low computational cost \cite{Mezard_GT},
although the BP algorithm does not achieve the information theoretic bound \cite{Johnson}.
Our contributions with regard to the BP algorithm are as follows:

\begin{description}
\item{(i)  } The true positive (TP) ratio, namely the ratio of infected patients reconstructed as positive, is improved to be greater than the TP probability in the test
by considering the confidence interval of the 
point estimate, corresponding to the infection probability.
A bootstrap method is introduced to construct the confidence interval. 

\item{(ii) } In our framework, prevalence, which is the fraction of the infected 
patients in the population, is introduced into the prior distribution
of the patients' state.
Prevalence is one of the fundamental measures in epidemiology,
and group testing has been applied to its estimation \cite{prevalence_Thompson,prevalence_Sobel,prevalence_Ron}.
The estimator of prevalence is constructed
in addition to the identification of the infected patients by
combining an expectation-maximization (EM) method with a BP algorithm.
Following the same procedure, the TP and FP probabilities of the 
test are estimated.

\item{(iii)} As another approach, the hierarchical Bayes model is introduced
to identify the infected patients and estimate the prevalence.
BP algorithm is applied to the hierarchical Bayes model and 
evaluate the performance in comparison with the approach described in (ii), and observe that
the computational cost of the hierarchical Bayes approach is lower than that 
required in (ii).

\end{description}

The remainder of the paper is organized as follows.
In Section \ref{sec:problem_setting},
the problem setting of group testing is explained
and introduce Bayesian inference.
In Section \ref{sec:BP},
the BP algorithm is introduced and 
show its reconstruction performance under finite 
false probabilities.
In Section \ref{sec:boot},
an estimator based on the 
confidence interval of the estimated infection probability is proposed.
In Section \ref{sec:EM},
the estimation of the unknown parameters in group testing is discussed by
using the likelihood calculated by the BP algorithm.
In Section \ref{sec:HBP},
the hierarchical Bayes model for group testing is introduced
and discuss the estimation of the prevalence by applying the BP algorithm
to the hierarchical model.
Section \ref{sec:summary} presents a summary and discussion.

\section{Problem setting}
\label{sec:problem_setting}

\begin{figure}
    \centering
    \includegraphics[width=4in]{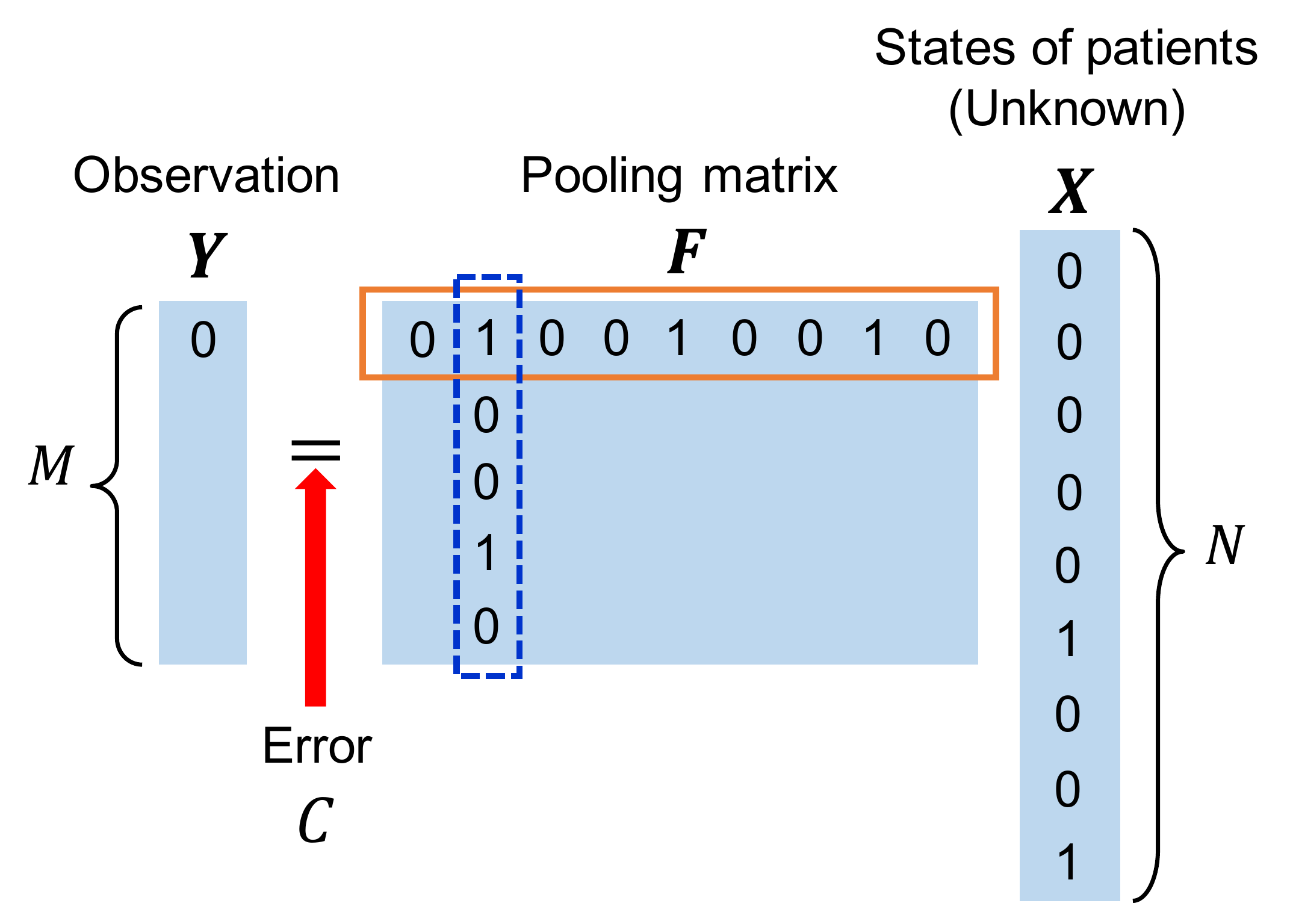}
    \caption{Matrix representation of group testing,
    where each pool size is $N_G=3$ and each overlap is $N_O=2$.
    The summation in the usual matrix product is replaced with a logical sum.}
    \label{fig:GT_matrix}
\end{figure}

We consider a population of $N$ patients and 
$M$ groups on which the test is performed.
Let us denote the state of $N$-patients by $\bm{X}^{(0)}\in\{0,1\}^N$,
where $X_i=1$ and $X_i=0$ 
means that $i$-th patient is infected and not infected,
respectively.
The grouping of the patients is determined by
the pooling matrix $\bm{F}\in\{0,1\}^{M\times N}$,
where $F_{\mu i}=1$ and $F_{\mu i}=0$
means that the $i$-th patient is in the $\mu$-th group
and not, respectively.
The true state of the $\mu~(=1,\cdots,M)$-th group, denoted by $Y_\mu^{(0)}$,
is given by
\begin{align}
    Y_\mu^{(0)}=\mathop{\vee}_{i=1}^NF_{\mu i}X_i^{(0)},
\end{align}
where $\vee_{i=1}^N f_i=f_1\vee f_2\vee \cdots\vee f_N$
denotes the logical sum of $N$ components.
Namely, when $\mu$-th pool contains 
at least one infected patient,
the state of the $\mu$-th pool is 1 (positive),
and 0 (negative) otherwise.

The test returns a false result with finite probability.
We assume that the errors in the tests are 
independent from each other 
and model the observation (result of the test) as 
\begin{align}
    Y_\mu = C(\mathop{\vee}_{i=1}^NF_{\mu i}X_i^{(0)}),
    \label{eq:Y}
\end{align}
where $C(\cdot)$ is the probabilistic function
whose behavior is given by \cite{Johnson}
\begin{align}
    P(C(a)=1|a=1)&=p_{\mathrm{TP}},~~~P(C(a)=0|a=1)=1-p_{\mathrm{TP}}\\
    P(C(a)=1|a=0)&=p_{\mathrm{FP}},~~~P(C(a)=0|a=0)=1-p_{\mathrm{FP}}.\label{eq:C_00}
\end{align}
Here, $p_{\mathrm{TP}}$ and $p_{\mathrm{FP}}$ correspond to the TP and FP probabilities
in the test, respectively,
and these values are common for all tests.
\Fref{fig:GT_matrix} shows matrix representation of the
group testing,
where the summations in the usual matrix factorization
are replaced with a logical sum.
We focus on the case $\alpha\equiv M\slash N<1$,
where the number of tests is smaller than that of the patients.
Hereafter, we consider that the size of group is fixed to $N_G$.
Further, the overlap, which is the 
number of groups that each patient belongs to,
is fixed at $N_O$;
hence, the relationship $N_O=\alpha\times N_G$ holds.

\subsection{Bayesian inference}

From the property of $C(\cdot)$,
the generative model of $\bm{Y}$
is given by
\begin{align}
    P(\bm{Y}&|\bm{X}^{(0)},p_{\mathrm{TP}},p_{\mathrm{FP}})=\prod_{\mu=1}^M P(Y_\mu|\bm{X}^{(0)},p_{\mathrm{TP}},p_{\mathrm{FP}})
    \label{eq:Y_generate_all}
\end{align}
where
\begin{align}
\nonumber
    P(Y_\mu &|\bm{X},p_{\mathrm{TP}},p_{\mathrm{FP}})=p_{\mathrm{TP}}
    Y_\mu T_\mu(\bm{X}^{(0)})+(1-p_{\mathrm{TP}})(1-Y_\mu)T_\mu(\bm{X}^{(0)})\\
    &+p_{\mathrm{FP}}Y_\mu(1-T_\mu(\bm{X}^{(0)}))+(1-p_{\mathrm{FP}})(1-Y_\mu)(1-T_\mu(\bm{X}^{(0)})),
    \label{eq:Y_generate}
\end{align}
and $T_\mu(\bm{X}^{(0)})=\mathop{\vee}_{i=1}^NF_{\mu i}X_i^{(0)}$.
The purpose is to infer the true states of patients $\bm{X}^{(0)}$
from the observation $\bm{Y}$.

In general, the true generative process of $\bm{Y}$ is unknown,
but it is reasonable to assume that the process is expressed by the
conditional Bernoulli distribution \eref{eq:Y_generate_all},
as the variables $\bm{Y}$ and $\bm{X}^{(0)}$ are binary,
although the value of 
true parameters $p_{\mathrm{TP}}$ and $p_{\mathrm{FP}}$ are not known in advance.
Bayesian inference is a preferred method in the presence of the
reasonable model.
As prior distribution of the patient states,
we use the following distribution:
\begin{align}
    P_0(\bm{X}|\rho)=\prod_{i=1}^N\{\rho X_i+(1-\rho)(1-X_i)\},
    \label{eq:prior}
\end{align}
where $\rho\in[0,1]$ is the prevalence, which is not known in advance.
The prevalence of the population does not 
necessarily match an 
individual's infection probability;
hence, the prior distribution \eref{eq:prior} is an assumption.
We consider that this form of prior information is appropriate 
for prevalence estimation 
as the prevalence of the population
generated by $\mathrm{Bernoulli}(\rho)$ converges to $\rho$
for sufficiently large $N$.

Following the Bayes rule, the
posterior distribution is given by
\begin{align}
    P(\bm{X}|\bm{Y})\propto P(\bm{Y}|\bm{X},\hat{p}_{\mathrm{TP}},\hat{p}_{\mathrm{FP}})P_0(\bm{X}|\hat{\rho}),
\end{align}
where $\hat{p}_{\mathrm{TP}}$, $\hat{p}_{\mathrm{TP}}$, and $\hat{\rho}$ 
are the assumed TP probability, FP probability,
and prevalence, respectively.
The $i$-th patient's state
is identified on the basis of the 
marginal distribution given by
\begin{align}
    P(X_i|\bm{Y})=\sum_{\bm{X}\backslash X_i}P(\bm{X}|\bm{Y})
\end{align}
where $\bm{X}\backslash X_i$ denotes the components of $\bm{X}$
other than $X_i$.
As the variable $X_i$ is binary, we can represent the 
marginal distribution using a Bernoulli
probability $\theta_i$ as
\begin{align}
    P(X_i|\bm{Y})=\theta_iX_i+(1-\theta_i)(1-X_i),
\end{align}
and $\theta_i$ corresponds to the infection probability,
namely, the probability that $X_i=1$.
The simplest estimate of $X_i^{(0)}$ is
the maximum a posteriori (MAP) estimator given by
\begin{align}
    X_i^{(\mathrm{MAP})}=\mathbb{I}(\theta_i>0.5),
\end{align}
where $\mathbb{I}(a)$ is the indicator function
whose value is 1 when $a$ is true, and 0 otherwise.

\section{Belief propagation}
\label{sec:BP}

\begin{figure}
    \centering
    \includegraphics[width=4in]{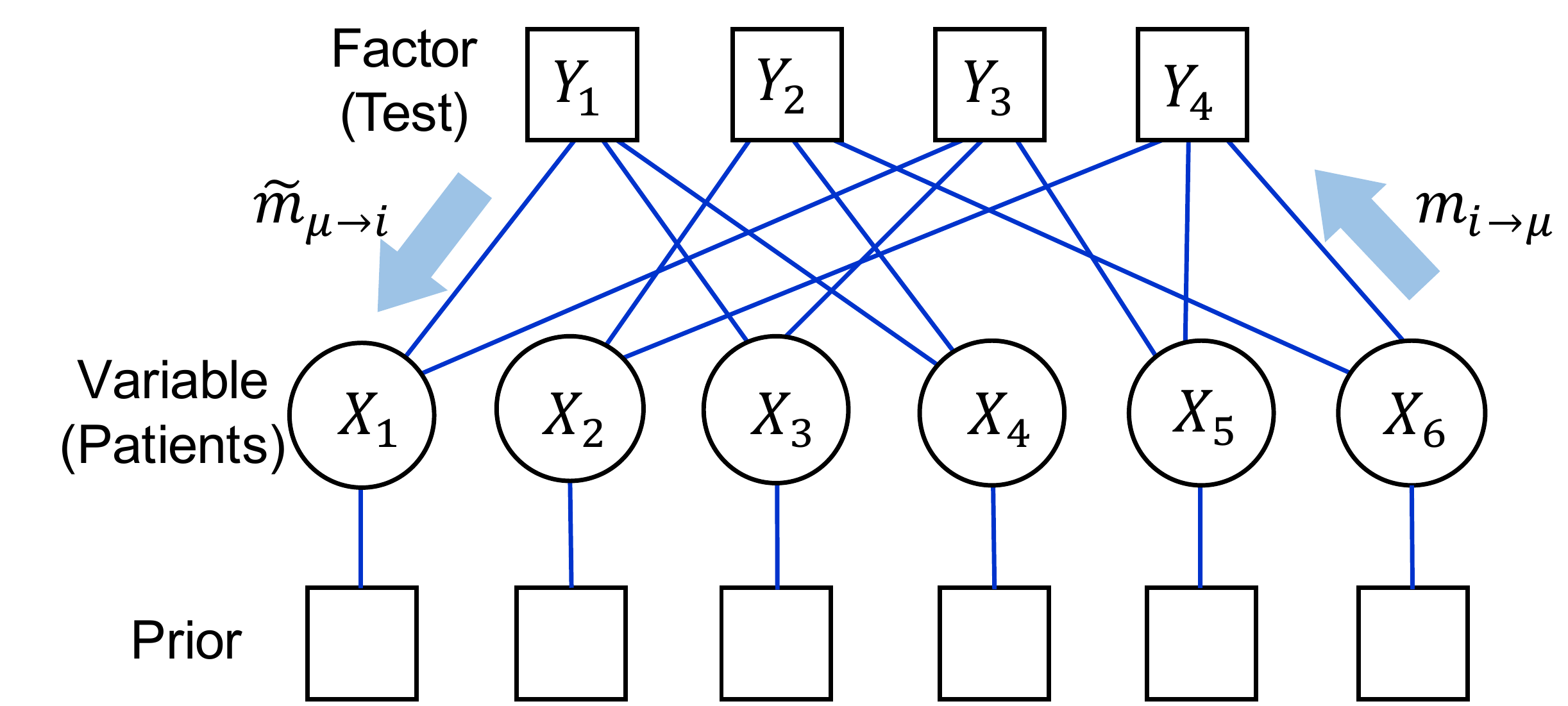}
    \caption{Factor graph representation of group testing for $N=6$, $M=4$,
    $N_G=3$, and $N_O=2$.}
    \label{fig:GT_graph}
\end{figure}

The computation of the marginal distribution
requires an exponential order of the sums,
and is thus intractable.
We approximately calculate the marginal distribution using 
the BP algorithm on the
factor graph representation of the group testing \cite{Johnson,Mezard-Montanari}.
Comparing the approximation by BP algorithm
with the exact calculation at a small size such as $N=20$,
we find that 
the accuracy of the BP algorithm is sufficient for applying 
it to the group testing problem (see Appendix\ref{sec:exact}).
As another numerical approach, Markov chain Monte Carlo (MCMC) method can be 
applied to the current problem setting;
however, MCMC requires high computational cost and is not feasible
particularly for estimating the unknown parameter, as discussed in Section \ref{sec:EM}.
In this study, we use the BP algorithm
as a reasonable method 
owing to its approximation accuracy and computational time.

\Fref{fig:GT_graph} shows the factor graph representation 
of the group testing for $N=6$, $M=4$, $N_G=3$, and $N_O=2$.
Here, ${\cal M}(\mu)$ and ${\cal G}(i)$
denote the indices of the patients in the $\mu$-th pool,
and those of the pools in which the $i$-th patient is included,
respectively.
The conditional probability $P(Y_\mu|\bm{X})$
depends on $X_i$~($i\in{\cal M}(\mu)$); hence,
the posterior distribution 
can be expressed as a bipartite graph, as shown in \Fref{fig:GT_graph}.
For the edge that connects the $\mu$-th factor (test)
and the $i$-th variable (patient), two types of messages are defined as
\begin{align}
    \tilde{m}_{\mu\to i}(X_i)&\propto\sum_{\bm{X}\backslash X_i}P(Y_\mu|\bm{X},\hat{p}_{\mathrm{TP}},\hat{p}_{\mathrm{FP}})\prod_{j\in{\cal M}(\mu)\backslash i}m_{j\to\mu}(X_i)\\
    m_{i\to\mu}(X_i)&\propto P_0(X_i|\hat{\rho})\prod_{\nu\in{\cal G}(i)\backslash\mu}\tilde{m}_{\nu\to i}(X_i),
\end{align}
which correspond to posterior information and output information, respectively.
Intuitively, the 
messages $m_{i\to \mu}(X_i)$ and $\tilde{m}_{\mu\to i}(X_i)$
represent the marginal distributions of $X_i$
before and after the $\mu$-th test is performed,
respectively.
As $X_i\in\{0,1\}$, these messages are represented 
by one parameter as
\begin{align}
    \tilde{m}_{\mu\to i}(X_i)&=\tilde{\theta}_{\mu\to i}X_i
    +(1-\tilde{\theta}_{\mu\to i})(1-X_i)\label{eq:one_para_h}\\
    m_{i\to\mu}(X_i)&=\theta_{i\to\mu}X_i+(1-\theta_{i\to\mu})(1-X_i),\label{eq:one_para}
\end{align}
where $\tilde{\theta}_{\mu\to i}$ and $\theta_{i\to\mu}$ are given by 
\begin{align}
    \tilde{\theta}_{\mu\to i}&=\frac{U_\mu}{\tilde{Z}_{\mu\to i}}
    \label{eq:hat_theta}\\
    \theta_{i\to\mu}&=\frac{\hat{\rho}\prod_{\nu\in{\cal G}(i)\backslash \mu} \tilde{\theta}_{\nu\to i}}{Z_{i\to\mu}}
    \label{eq:theta_h}
\end{align}
and
\begin{align}
    U_\mu&=p_{\mathrm{TP}}Y_\mu+(1-p_{\mathrm{TP}})(1-Y_\mu)\label{eq:U}\\
    W_\mu&=p_{\mathrm{FP}}Y_\mu+(1-p_{\mathrm{FP}})(1-Y_\mu)\label{eq:W}
\end{align}
\begin{align}
    \tilde{Z}_{\mu\to i}&=U_\mu\left(2-\prod_{j\in{\cal M}(\mu)\backslash i} (1-\theta_{j\to\mu})\right)
    +W_\mu\prod_{j\in{\cal M}(\mu)\backslash i}(1-\theta_{j\to\mu})\\
    Z_{i\to\mu}&=\hat{\rho}\prod_{\nu\in{\cal G}(i)\backslash \mu} \tilde{\theta}_{\nu\to i}+(1-\hat{\rho})\prod_{\nu\in{\cal G}(i)\backslash \mu} (1-\tilde{\theta}_{\nu\to i}).
\end{align}
Using these messages, we can approximate the marginal distribution as 
\begin{align}
\nonumber
    P(X_i)&\propto \{\hat{\rho}X_i+(1-\hat{\rho})(1-X_i)\}\prod_{\mu\in{\cal G}(i)}\tilde{m}_{\mu\to i}(X_i)
    \\
    &=\left(\hat{\rho}\!\!\!\prod_{\mu\in{\cal G}(i)}\!\!\tilde{\theta}_{\mu\to i}\right)X_i+
    \left((1-\hat{\rho})\!\!\prod_{\mu\in{\cal G}(i)}\!\!(1-\tilde{\theta}_{\mu\to i})\right)(1-X_i),
\end{align}
and thus the infection probability is approximated as 
\begin{align}
    \hat{\theta}_i=\frac{\hat{\rho}\prod_{\mu\in{\cal G}(i)}\tilde{\theta}_{\mu\to i}}
    {\hat{\rho}\prod_{\mu\in{\cal G}(i)}\tilde{\theta}_{\mu\to i}
    +(1-\hat{\rho})\prod_{\mu\in{\cal G}(i)}(1-\tilde{\theta}_{\mu\to i})},
    \label{eq:infect_prob}
\end{align}
and the MAP estimator is given by
\begin{align}
\hat{X}_i^{(MAP)}=\mathbb{I}(\hat{\theta}_i>0.5).
\end{align}

\begin{algorithm}[H]
\caption{BP for Bayesian Group Testing}
\label{alg:BP}
\begin{algorithmic}[1]
\Require {$\bm{Y}\sim P(\bm{Y}|\bm{X}^{(0)})$ and $\bm{F}$}
\Ensure {$\bm{\theta}\in[0,1]^N$}
    \State{$\{\theta^{(0)}_{i\to\mu}\}\gets$ initial values from $[0,1]^{N\times M}$}
    \State{$\{\hat{\theta}^{(0)}_{\mu\to i}\}\gets$ initial values from $[0,1]^{M\times N}$}
    \State{$\bm{U}\gets~p_{\mathrm{TP}}\bm{Y}+(1-p_{\mathrm{TP}})(\bm{1}_M-\bm{Y})$}
    \State{$\bm{W}\gets~p_{\mathrm{FP}}\bm{Y}+(1-p_{\mathrm{FP}})(\bm{1}_M-\bm{Y})$}
    \For{$t = 1 \, \ldots \, T$}
        \For {all combinations of $(\mu,i)$ such that $F_{\mu i}=1$}
            \State{$\tilde{Z}_{\mu\to i}^{(t)}\gets U_\mu\left\{2-\prod_{j\in{\cal M}(\mu)\backslash i}\left(1-\theta_{j\to\mu}^{(t-1)}\right)\right\}
    +W_\mu\prod_{j\in{\cal M}(\mu)\backslash i}\left(1-\theta_{j\to\mu}^{(t-1)}\right)$}
            \State{$Z_{i\to\mu}^{(t)}\gets\rho \prod_{\nu\in{\cal G}(i)\backslash \mu} \tilde{\theta}_{\nu\to i}^{(t-1)}+(1-\rho)\prod_{\nu\in{\cal G}(i)\backslash \mu} \left(1-\tilde{\theta}_{\nu\to i}^{(t-1)}\right)$}
            \State{$\tilde{\theta}_{\mu\to i}^{(t)}\gets\frac{U_\mu}{\tilde{Z}^{(t)}_{\mu\to i}}$}
            \State{$\theta_{i\to\mu}^{(t)}\gets\frac{\rho \prod_{\nu\in{\cal G}(i)\backslash \mu} \tilde{\theta}_{\nu\to i}^{(t-1)}}{Z^{(t)}_{i\to\mu}}$}
        \EndFor
    \EndFor
    \For{$i = 1 \, \ldots \, N$}
        \State{$\hat{\theta}_i\gets$ $\frac{\rho\prod_{\mu\in{\cal G}(i)}\tilde{\theta}_{\mu\to i}^{(T)}}
    {\rho\prod_{\mu\in{\cal G}(i)}\tilde{\theta}_{\mu\to i}^{(T)}
    +(1-\rho)\prod_{\mu\in{\cal G}(i)}\left(1-\tilde{\theta}_{\mu\to i}^{(T)}\right)}$}
    \EndFor
\end{algorithmic}
\end{algorithm}

First, 
we consider the case where 
we know the correct parameters; 
$\hat{p}_{\mathrm{TP}}=p_{\mathrm{TP}},~\hat{p}_{\mathrm{FP}}=p_{\mathrm{FP}}$, and $\hat{\rho}=\rho$.
The pseudocode of the BP algorithm for group testing with known parameters
is shown in Algorithm \ref{alg:BP}.
We introduce a damping factor $d\in(0,1]$ for the stabilization of the algorithm as
\begin{align}
\tilde{\theta}_{\mu\to i}^{(t)}
&\gets d\tilde{\theta}_{\mu\to i}^{(t)}+(1-d)\tilde{\theta}_{\mu\to i}^{(t-1)}\\
\theta_{i\to\mu}^{(t)}&\gets d\theta_{i\to\mu}^{(t)}+(1-d)\theta_{i\to\mu}^{(t-1)}.
\end{align}
In all the numerical simulations performed in this study, 
we set $d=0.1$ conservatively; however, this choice extends the 
time for convergence.
An adaptive setting of the damping factor for group testing is worth studying to achieve faster and stable convergence \cite{Schniter}.

\begin{figure}
\begin{minipage}{0.49\hsize}
\centering
    \includegraphics[width=2.7in]{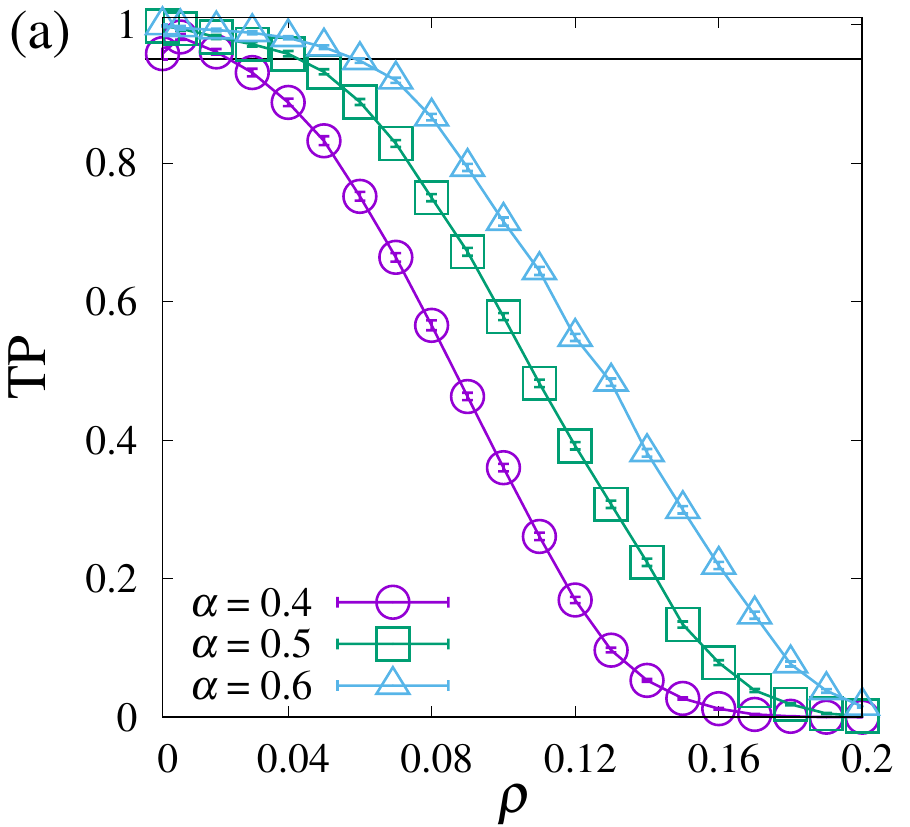}
\end{minipage}
\begin{minipage}{0.49\hsize}
\centering
\includegraphics[width=2.7in]{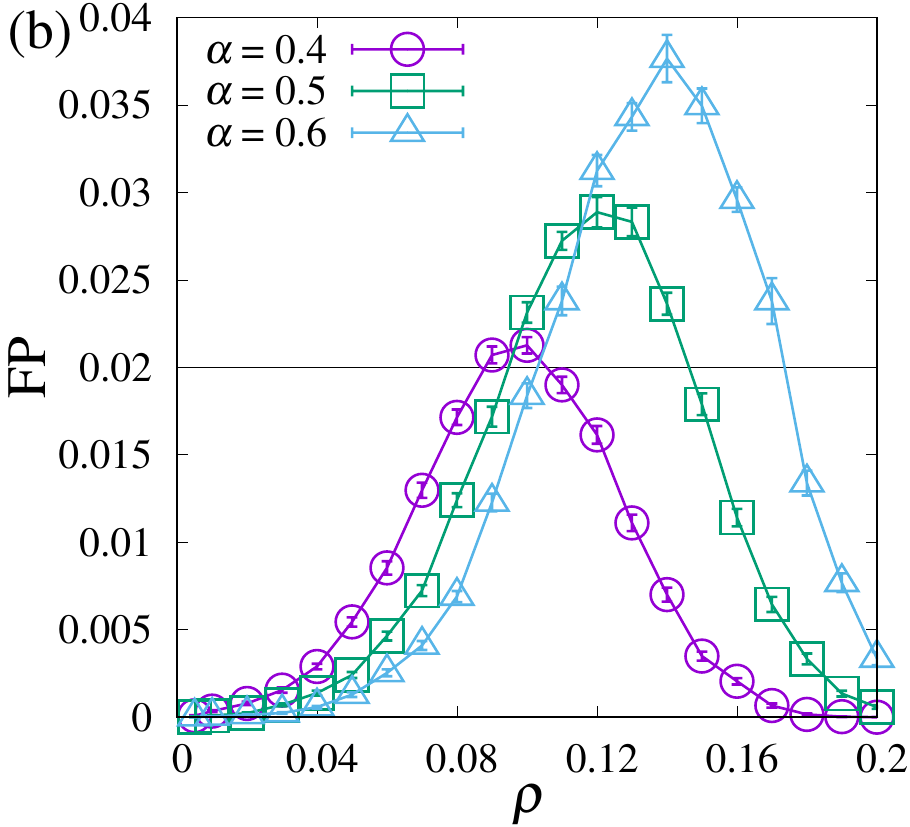}
\end{minipage}
    \caption{$\rho$-dependence of (a) TP and (b) FP at $N=1000$ and 
    $N_G=10$ for various $\alpha=M\slash N$. 
    Error probabilities on the test are fixed at $p_{\mathrm{TP}}=0.95$ and $p_{\mathrm{FP}}=0.02$.
    The horizontal lines in (a) and (b) represent $p_{\mathrm{TP}}$ and 
    $p_{\mathrm{FP}}$, respectively.}
    \label{fig:w0d02}
\end{figure}

\begin{figure}
\begin{minipage}{0.49\hsize}
\centering
    \includegraphics[width=2.7in]{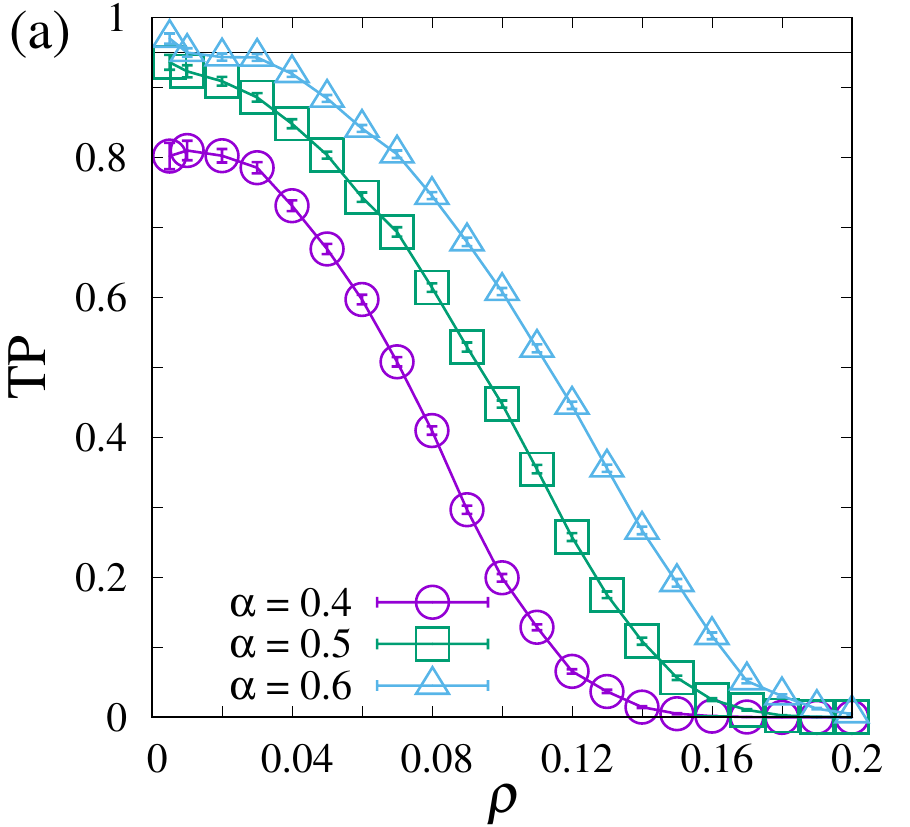}
\end{minipage}
\begin{minipage}{0.49\hsize}
\centering
\includegraphics[width=2.7in]{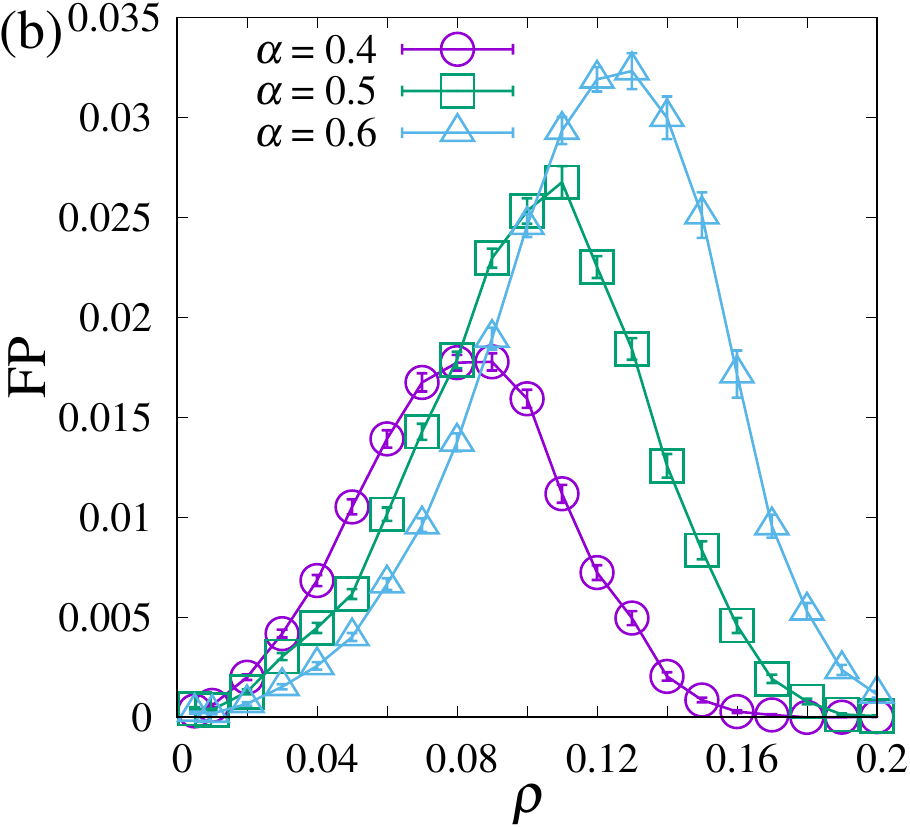}
\end{minipage}
    \caption{$\rho$-dependence of (a) TP and (b) FP at $N=1000$ and 
    $N_G=10$ for various $\alpha$. Error rates are fixed at $p_{\mathrm{TP
    }}=0.95$ and $p_{\mathrm{FP}}=0.1$.
    The horizontal line in (a) represents $p_{\mathrm{TP}}$.
    The FP region shown in (b) is below $p_{\mathrm{FP}}$.}
    \label{fig:w0d1}
\end{figure}

We check the performance of BP algorithm for 
randomly constructed pooling matrix 
under the constraint 
as $\sum_iF_{\mu i}=N_G~\forall \mu$ and 
$\sum_\mu F_{\mu i}=N_O~\forall i$.
The true state of patients $\bm{X}^{(0)}$ is also randomly generated
under the constraint $\sum_iX_i^{(0)}=N\rho$.
The accuracy of the MAP estimator is measured using 
the TP rate and FP rate, 
given by
\begin{align}
    \mathrm{TP}&=\frac{\frac{1}{N}\sum_i X_i^{(0)}\hat{X}_i^{(MAP)}}{\frac{1}{N}\sum_i X_i^{(0)}}\\
    \mathrm{FP}&=\frac{\frac{1}{N}\sum_i(1-X_i^{(0)})\hat{X}^{(MAP)}_i}{1-\frac{1}{N}\sum_i X_i^{(0)}},
\end{align}
respectively.
A TP value larger than $p_{\mathrm{TP}}$ and an FP value smaller than 
$p_{\mathrm{FP}}$ indicate that 
the performance of the BP-based identification is
better than the parallel test of $N$-patients.
\Fref{fig:w0d02} shows $\rho$-dependence of (a) TP
and (b) FP at $N=1000$, $N_G=10$, $p_{\mathrm{TP}}=0.95$, and 
$p_{\mathrm{FP}}=0.02$, respectively.
Each data point represents the averaged value with respect to 
100 realizations of $\bm{Y}$ and $\bm{X}^{(0)}$.
The horizontal lines in (a) and (b) indicate 0.95 and 0.02, which 
are the TP and FP probabilities of the test, respectively.
As $\alpha$ increases, that is, as the number of tests increases,
TP increases and the $\rho$ region where TP is larger than $p_{\mathrm{TP}}$ extends.
FP also has a smaller value than $p_{\mathrm{FP}}$ for small values of $\rho$,
and this success region extends as $\alpha$ increases.
A similar tendency is shown for different values of $p_{\mathrm{TP}}$ and $p_{\mathrm{FP}}$.
As an example, we show the TP and FP for $N=1000$, $N_G=10$, $p_{\mathrm{TP}}=0.95$, and $p_{\mathrm{FP}}=0.1$
in \Fref{fig:w0d1}.
The FP probability $p_{\mathrm{FP}}$ has significant influence on TP,
which is evident from the comparison of Figs. \ref{fig:w0d02} and \ref{fig:w0d1}.
Intuitively, an
increase in $p_{\mathrm{FP}}$ (or decrease in $p_{\mathrm{TP}}$)
causes uncertainty of the identification and 
decreases $\hat{\theta}_i$; hence,
the MAP estimator tends to be zero.

\begin{figure}
\begin{minipage}{0.49\hsize}
\centering
    \includegraphics[width=2.7in]{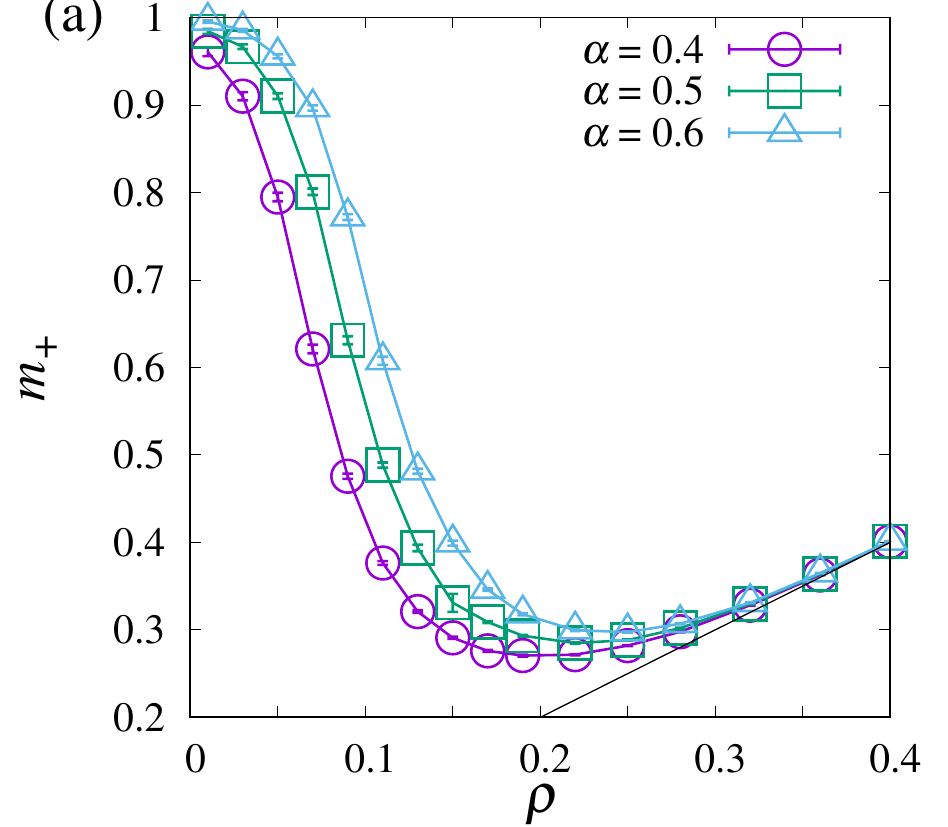}
\end{minipage}
\begin{minipage}{0.49\hsize}
\centering
\includegraphics[width=2.7in]{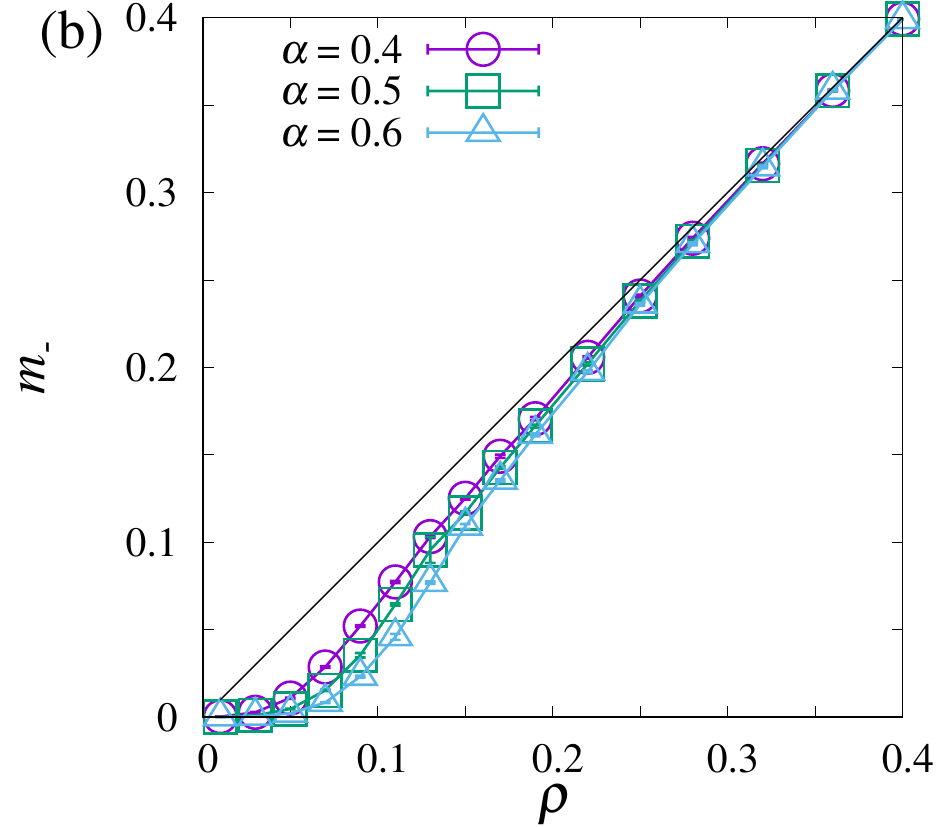}
\end{minipage}
    \caption{$\rho$-dependence of (a) $m_+$ and (b) $m_-$ at $N=1000$ and 
    $N_G=10$ for various $\alpha$. Error rates are fixed at $p_{\mathrm{TP
    }}=0.95$ and $p_{\mathrm{FP}}=0.02$.
    The gradients of solid lines are 1.}
    \label{fig:magnetization}
\end{figure}

As shown in Figs. \ref{fig:w0d02} and \ref{fig:w0d1},
the TP and FP converge to zero as $\rho$ increases.
To understand this behavior,
let us consider the following quantities
\begin{align}
m_+&=\frac{1}{N\rho}\sum_{i=1}^NX_i^{(0)}\hat{\theta}_i\\
m_-&=\frac{1}{N(1-\rho)}\sum_{i=1}^N(1-X_i^{(0)})\hat{\theta}_i,
\end{align}
which correspond to the magnetization for infected patients and 
non-infected patients, respectively.
\Fref{fig:magnetization} shows the $\rho$-dependence of 
(a) $m_+$ and (b) $m_-$ at $N=1000$, $N_G=10$, $p_{\mathrm{TP}}=0.95$,
and $p_{\mathrm{FP}}=0.02$, which are the same parameters as those used for
\Fref{fig:w0d02}.
The values $m_+$ and $m_-$ 
converge to $\rho$ from above and from below, respectively,
as $\rho$ increases.
This means that the infection probability $\hat{\theta}_i$
tends to be $\rho$ without depending on the value $X_i^{(0)}$
at a sufficiently large $\rho$.
Therefore, from the definition of the MAP estimator,
$\hat{X}_i^{(\mathrm{MAP})}=0$ holds for any $i$,
resulting in $\mathrm{TP}=\mathrm{FP}=0$,
when $\rho(<0.5)$ is sufficiently large
to be $m_\to\rho$ and $m_-\to\rho$.
At $\rho\geq 0.5$, the estimate becomes
$\hat{\theta}_i\to \rho\geq 0.5$ for any $i$;
hence, $\hat{X}_i^{(\mathrm{MAP})}=1$ holds and $\mathrm{TP}=\mathrm{FP}=1$.

I note that the relationship $\hat{\theta}_i\to \rho$
is caused by the modeling and not the BP algorithm.
In fact, the exact calculation of the posterior distribution
and the corresponding MAP estimator achieve similar result,
as shown in Appendix\ref{sec:exact}.
In the BP algorithm, $\hat{\theta}_i\to\rho$ is achieved by
messages $\tilde{\theta}_{\mu\to i}\to 0.5$
and $\theta_{i\to\mu}\to\rho$ for any $\mu$ such that $F_{\mu i}=1$.
These messages indicate that 
the infection probability 
is determined by the prior distribution.

\begin{figure}
\begin{minipage}{0.495\hsize}
\centering
    \includegraphics[width=2.7in]{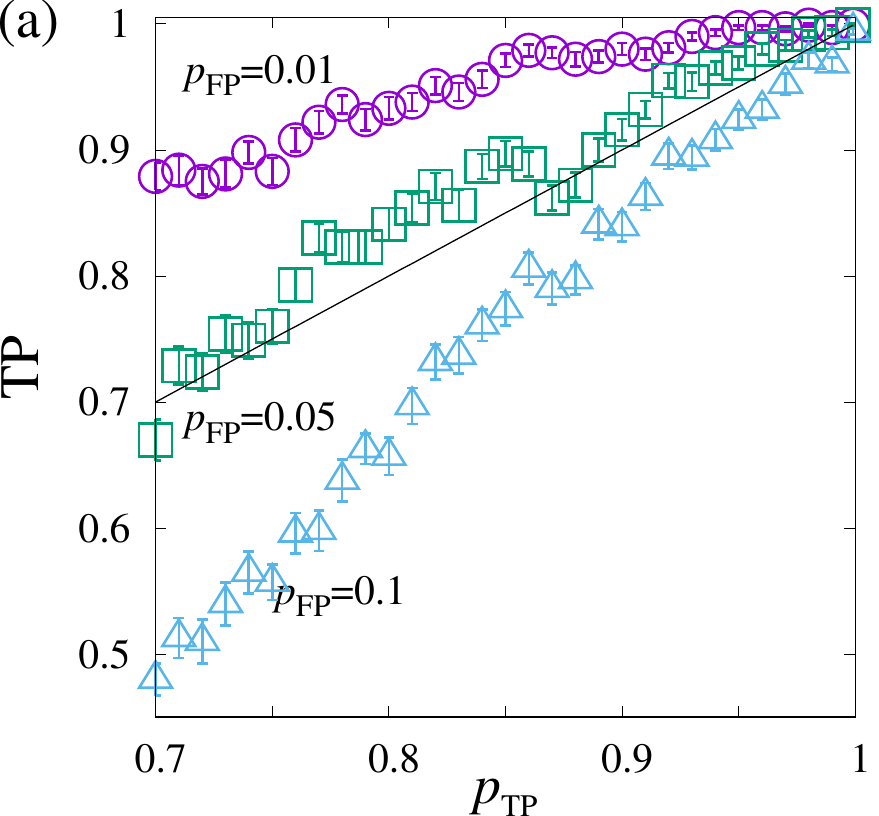}
\end{minipage}
\begin{minipage}{0.495\hsize}
\centering
\includegraphics[width=2.7in]{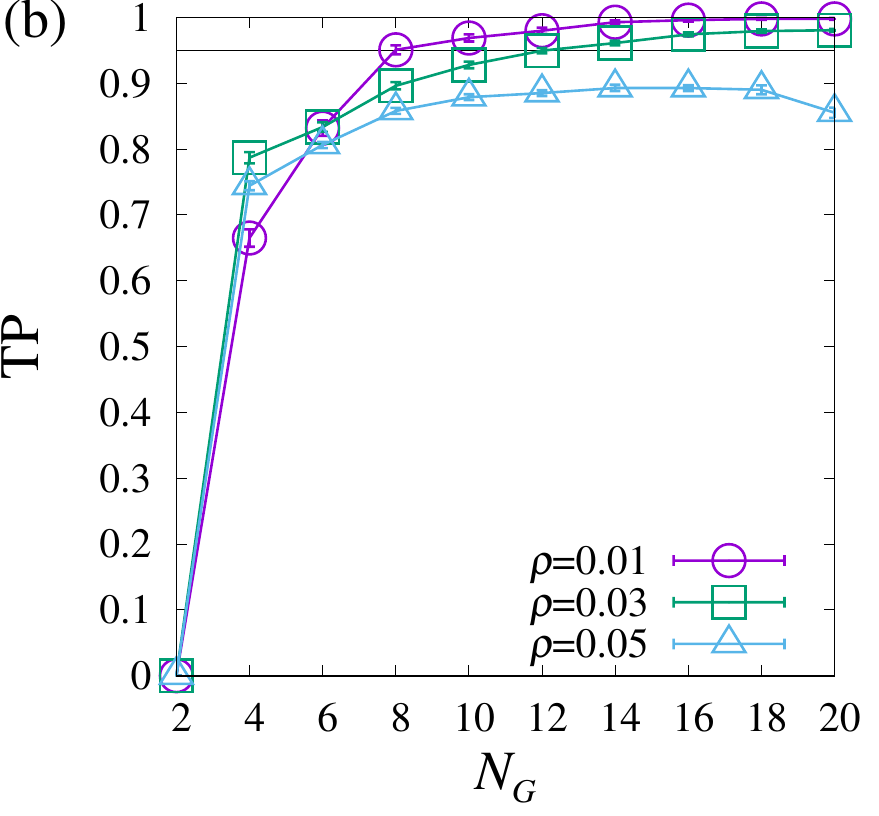}
\end{minipage}
    \caption{(a) $p_{\mathrm{TP}}$-dependence of TP at $N=1000, M=500, N_G=10~(N_O=5)$, and $\rho=0.01$
    for different values of $p_{\mathrm{FP}}$.
    (b) $N_G$-dependence of TP at $N=1000, M=500$ and $p_{\mathrm{TP}}=0.95, p_{\mathrm{FP}}=0.05$.
    The solid line indicates $0.95~(p_{\mathrm{TP}})$.
    }
    \label{fig:TP_vs_u_M500}
\end{figure}

The dependence of the TP on $p_{\mathrm{TP}}$ and $p_{\mathrm{FP}}$
is shown in \Fref{fig:TP_vs_u_M500}(a)
at $N=1000$, $N_G=10$ ($N_O=5$), and $\rho=0.01$.
The solid line indicates $\mathrm{TP}=p_{\mathrm{TP}}$, and
a TP value over the solid line means that the
reconstruction by the BP algorithm achieves a higher TP
than the parallel test of $N$-patients,
and this situation is achieved at a
sufficiently small FP probability $p_{\mathrm{FP}}<0.05$
in this parameter region.
The reconstruction performance also depends on $N_G$.
\Fref{fig:TP_vs_u_M500} (b) 
shows the $N_G$-dependence of the TP
at $N=1000$, $M=500$, $p_{\mathrm{TP}}=0.95$, and $p_{\mathrm{FP}}=0.05$.
The TP increases as $N_G$ increases without increasing the number of tests.

In practical testing,
one of the objectives is
to identify the infected patients
to prevent the spread of the disease.
Therefore, increasing the TP is a priority,
and we mainly focus on the improvement of the TP
using the BP algorithm.

\subsection{BP algorithm needs ``decision threshold''}

Before proceeding to the improvement of the TP,
we discuss the trivial fixed points of the BP algorithm
and introduce the idea of ``decision threshold'' for the identification of 
infected patients.

From \eref{eq:infect_prob}, when
$\tilde{\theta}_{\mu\to i}=1$ for $\mu\in{\cal G}(i)$, we obtain $\hat{\theta}_i=1$ 
irrespective of the value of $\hat{\rho}$.
This situation arises when 
$\theta_{j\to\mu}=0$ for $j\in{\cal M}(\mu)\backslash i$ at 
$p_{\mathrm{FP}}=0$.
Therefore, $\hat{\theta}_i=1$ is achieved when 
the patients $j\in{\cal M}(\mu)\backslash i$
are estimated as negative before $\mu$-th test is performed, where $\mu\in{\cal G}(i)$.
This is the case in which the $i$-th patient is trivially 
identified as positive.
In other words, the BP algorithm does not return $\hat{\theta}_i=1$
for general cases; hence,
to determine the infected patients $X_i^{(0)}\in\{0,1\}$ from 
an estimate at the BP fixed point $\hat{\theta}_i\in[0,1]$,
we need a ``decision threshold'' such as a MAP estimator,
where $\hat{\theta}_i=0.5$ is the threshold for determining the
infected patients.
The TP and FP depend on this threshold,
and our strategy for the improvement of TP is an 
appropriate threshold choice, as discussed in the next section.

The threshold at $\hat{\theta}_i=0$ is expected to obtain a conservative
result; however, it is not appropriate for a general value of $p_{\mathrm{TP}}$.
Following a similar logic,
$\hat{\theta}_i=0$ is obtained when 
at least one of the components of $\tilde{\theta}_{\mu\to i}$ among 
$\mu\in{\cal G}(i)$ takes the value 0, which is achieved at $p_{\mathrm{TP}}=1$ and $Y_\mu=0$
or $p_{\mathrm{TP}}=0$ and $Y_\mu=1$.
The former case means that 
all the patients belonging to the $\mu$-th test 
are negative when $Y_\mu=0$ and $p_{\mathrm{TP}}=1$.
In the latter case,
$Y_\mu=1$ means $Y_\mu^{(0)}=0$ because $p_{\mathrm{TP}}=0$;
Hence, all the patients belonging to the positive test
are negative.
In other words, 
$\hat{\theta}_i$ is always larger than zero
when $p_{\mathrm{TP}}$ is less than 1 or more than 1.
Therefore, all the patients are determined as positive 
under the threshold at $\hat{\theta}_i=0$,
which corresponds to $\mathrm{FP}=1$.

\section{Improvement of true-positive rate considering fluctuation of the estimates}
\label{sec:boot}

The estimated Bernoulli probability $\hat{\bm{\theta}}$
is a function of $\bm{Y}$,
and fluctuates depending on the probabilistic observation.
The quantification of the credibility
of $\hat{\bm{\theta}}$ helps in determining 
the infected patients under conditions of noisy observation data.
The confidence interval is one of the guides
in inference considering the input fluctuation \cite{Sparse_book}.
For convenience, we introduce the following statistic:
\begin{align}
    \hat{\tau}_i\equiv\log\frac{\hat{\theta}_i}{1-\hat{\theta}_i},
    \label{eq:tau_def}
\end{align}
which gives the MAP estimator as 
\begin{align}
    \hat{X}_i^{\mathrm{MAP}}=\mathbb{I}(\hat{\tau}_i>0).
    \label{eq:MAP_tau}
\end{align}
Here, we assume that the generative model has the corresponding ``true value'' $\tau_i$.
Following the normal theory, the
$95\%$ confidence interval of the true value of $\tau_i$ is constructed as
\begin{align}
   \tau_i\in[\hat{\tau}_i-1.96\hat{\sigma}_i,\hat{\tau}_i+1.96\hat{\sigma}_i],
   \label{eq:conf_int}
\end{align}
where $1.96$ is the $97.5\%$ quantile of the standard normal distribution,
and $\hat{\sigma}_i$ is the estimate of the standard error.
We resort to the nonparametric bootstrap method 
to estimate the standard error \cite{Efron}.
We generate $b=1,\cdots,N_B$ bootstrap samples 
$\bm{Y}^{(b)}\in\{0,1\}^M$
and $\bm{F}^{(b)}\in\{0,1\}^{M\times N}$ as
\begin{align}
    \{y_\mu^{(b)},\tilde{\bm{F}}_\mu^{(b)}\}\sim \hat{P}(y,\tilde{\bm{F}}),
\end{align}
where $\tilde{\bm{F}}_\mu^{(b)}$
is the $\mu$-th row vector of $\bm{F}^{(b)}$
and $\hat{P}(Y,\tilde{\bm{F}})$ is the empirical distribution
of given $\bm{Y}$ and $\bm{F}$ defined by
\begin{align}
    \hat{P}(Y,\tilde{\bm{F}})=\frac{1}{M}\sum_{\nu=1}^M\delta(y-y_\nu)\delta(\tilde{\bm{F}}-\tilde{\bm{F}}_\nu).
\end{align}
For the construction of the confidence interval for the Bayesian point estimate,
the parametric bootstrap method is another approach \cite{Hjort_Bayes_boot,Efron_Bayes_boot}
where the bootstrap samples are generated according to the posterior distribution
with the point estimate.
Here, we use the nonparametric approach for its simplicity.

We execute the BP algorithm for every bootstrap sample,
and denote the estimate under the $b$-th bootstrap sample as $\bm{\hat{\tau}}^{(b)}$.
The bootstrap sample contains the same row vector of $\bm{F}$ with high probability.
We omit the overlapped rows 
to stabilize the BP algorithm.
Using $\tau_i^{(b)}~(b=1,\cdots,N_B)$,
we obtain the estimate of the standard error of $\hat{\theta}_i$ as
\begin{align}
    \hat{\sigma}_i=\sqrt{\frac{1}{N_B-1}\sum_{b=1}^{N_B}(\hat{\tau}_i^{(b)}-\overline{\tau_i})^2},
\end{align}
where $\overline{\tau}_i\equiv\frac{1}{N_B}\sum_{b=1}^{N_B}\hat{\tau}_i^{(b)}$
is the average over the bootstrap samples.

\begin{figure}
    \begin{minipage}{0.495\hsize}
    \centering
    \includegraphics[width=2.7in]{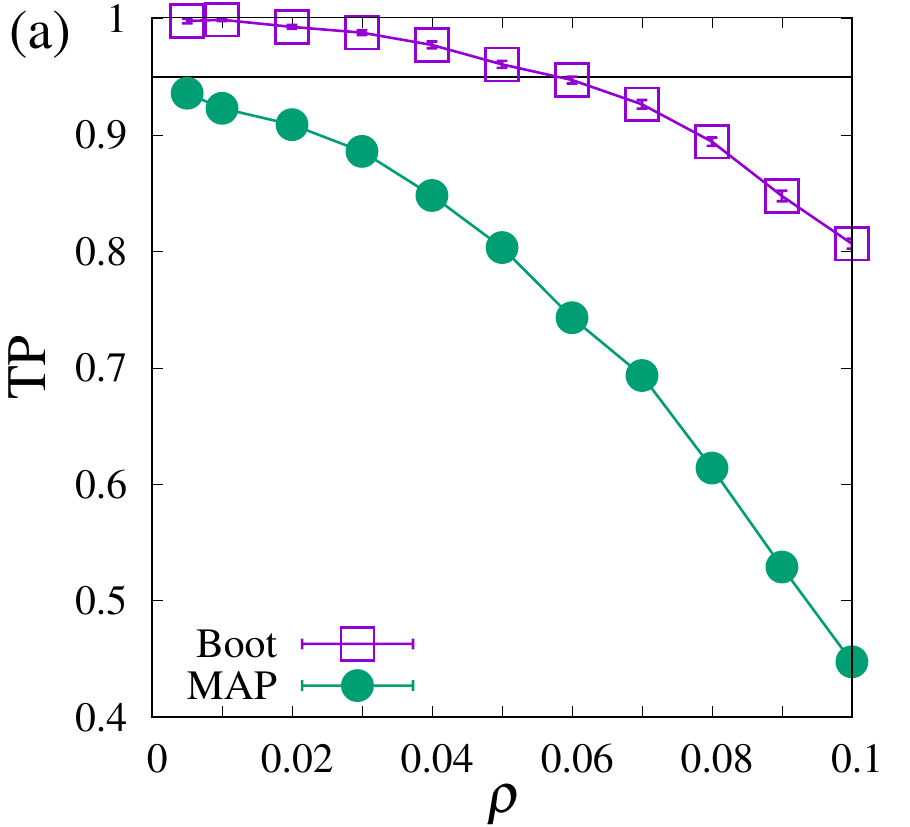}
    \end{minipage}
    \begin{minipage}{0.495\hsize}
    \centering
    \includegraphics[width=2.7in]{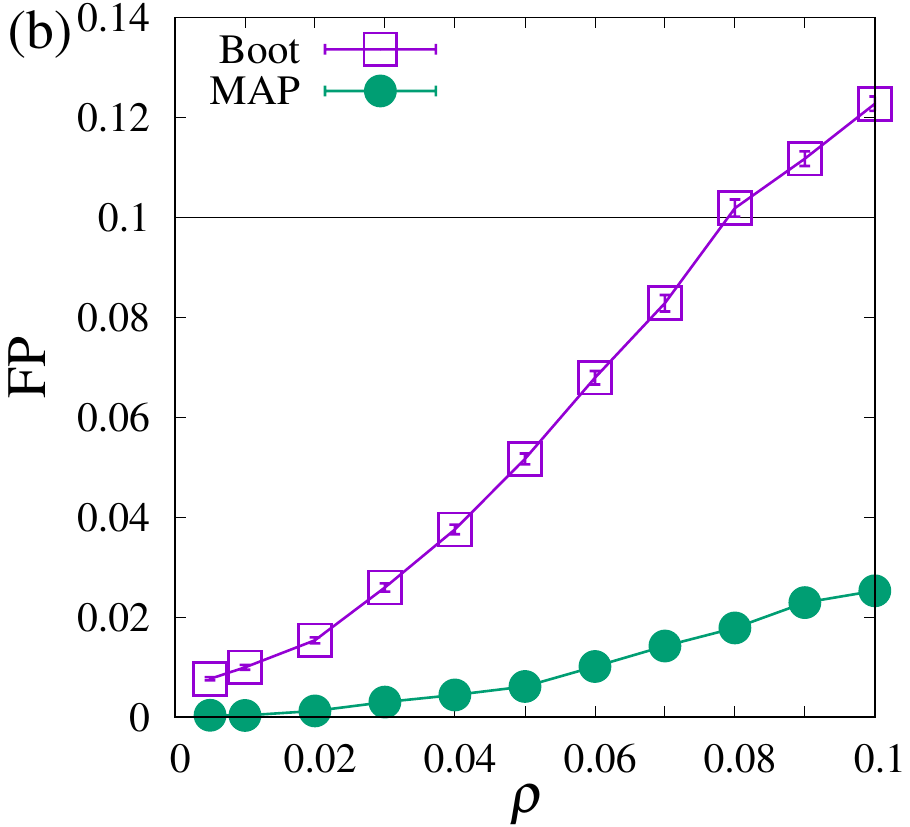}
    \end{minipage}
    \caption{$\rho$-dependence of 
    (a) TP and (b) FP bootstrap estimator for $N=1000$, $M=500$, $N_G=10$, $p_{\mathrm{TP}}=0.95$, and $p_{\mathrm{FP}}=0.1$.}
    \label{fig:Boot_N1000_M500}
\end{figure}

\begin{figure}
    \begin{minipage}{0.495\hsize}
    \centering
    \includegraphics[width=2.7in]{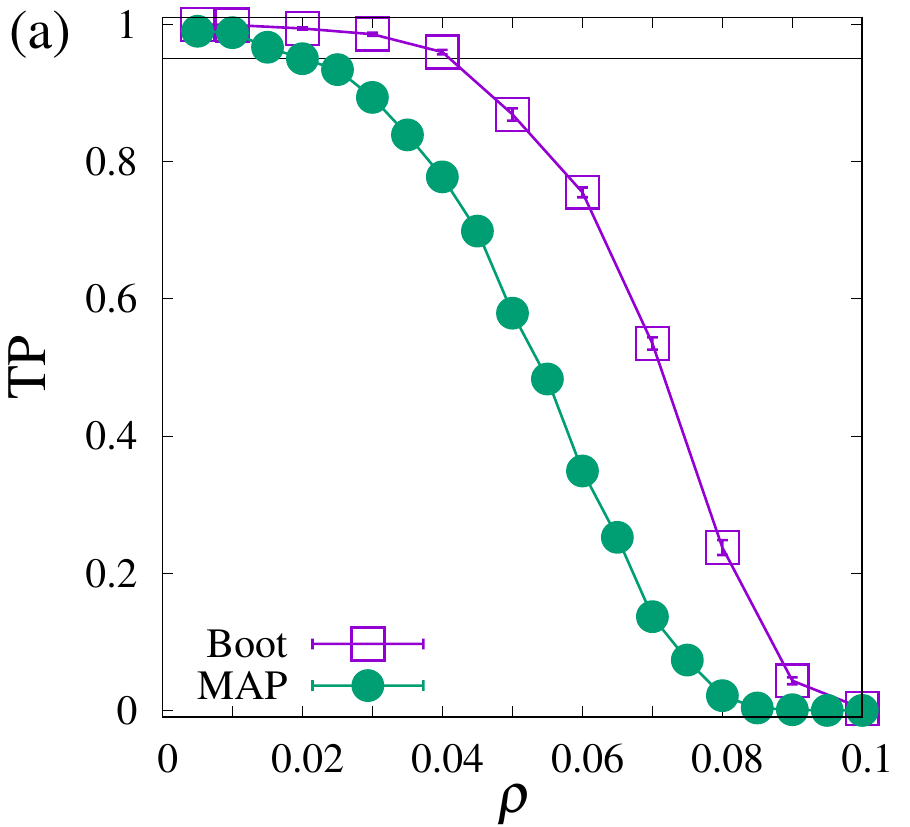}
    \end{minipage}
    \begin{minipage}{0.495\hsize}
    \centering
    \includegraphics[width=2.7in]{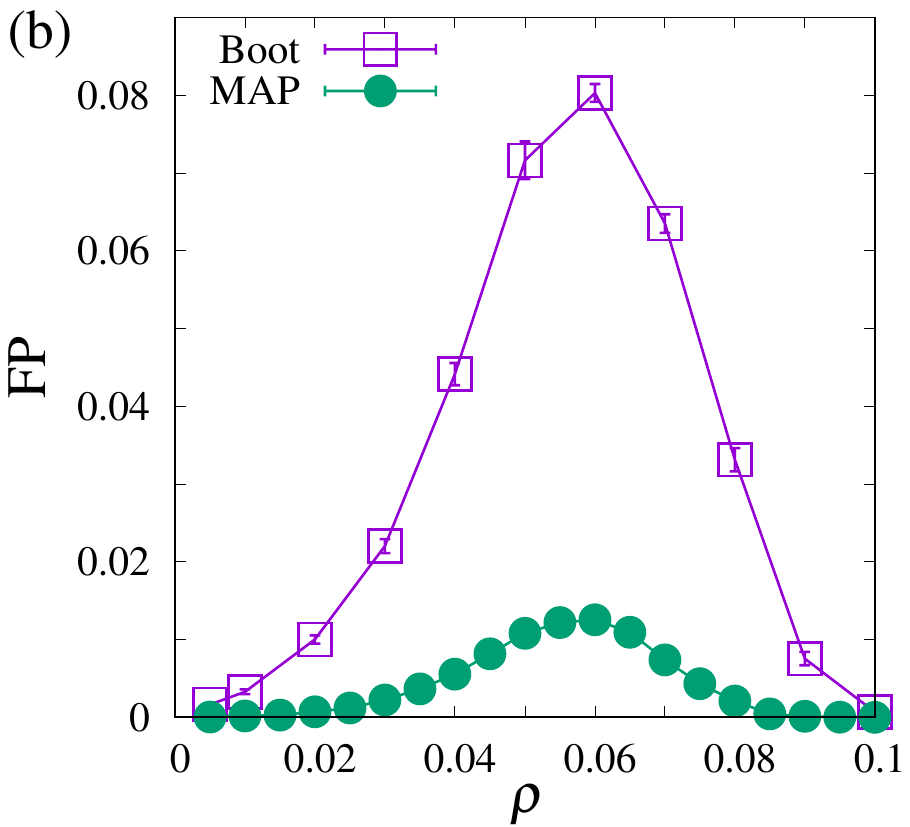}
    \end{minipage}
    \caption{$\rho$-dependence of 
    (a) TP and (b) FP bootstrap estimator for $N=1000$, $M=400$, $N_G=20$, $p_{\mathrm{TP}}=0.95$, and $p_{\mathrm{FP}}=0.1$.}
    \label{fig:Boot_N1000_M400}
\end{figure}

We define the bootstrap estimate of the $i$-th patient's state as 
\begin{align}
    \hat{X}_i^{\mathrm{(Boot)}}=\mathrm{I}(\hat{\tau}_i+1.96\hat{\sigma}_i>0),
    \label{eq:boot_estimator}
\end{align}
indicating that the patients whose confidence interval
runs over the region $\tau>0$ are regarded as infected.
In comparison with the MAP estimator \eref{eq:MAP_tau},
the decision threshold over which the patients are estimated as infected is lower by $1.96\hat{\sigma}_i$.
Further, when $\hat{X}^{\mathrm{(MAP)}}=1$, $\hat{X}^{\mathrm{(Boot)}}=1$ always.
Hence, \eref{eq:boot_estimator} can change the results of the patients
who are determined to be non-infected using the MAP estimator.
\Fref{fig:Boot_N1000_M500} shows the 
(a) TP and (b) FP of the bootstrap estimate
at $N=1000$, $M=500$, $N_G=10$, $p_{\mathrm{TP}}=0.95$, and $p_{\mathrm{FP}}=0.1$.
We generate $N_B=1000$ bootstrap samples for each set of sample 
$\{\bm{Y},\bm{F},\bm{X}^{(0)}\}$, and each point is averaged over 100 samples.
The MAP estimator cannot achieve a higher TP than $p_{\mathrm{TP}}$ for any $\rho$;
however, the bootstrap estimator improves the TP to be greater than $p_{\mathrm{TP}}$.
Moreover, the FP of the bootstrap estimator is higher than that of the MAP estimator;
this is caused by the reduced decision threshold compared with that of the MAP estimator.
However, the FP is smaller than $p_{\mathrm{FP}}$ for a sufficiently small $\rho$; hence, 
I consider the 
bootstrap estimator as practicable.
The situation is the same for other parameter regions.
As an example, we show the TP and FP of bootstrap estimator
at $N=1000$, $M=400$, $N_G=20$, $p_{\mathrm{TP}}=0.95$, and $p_{\mathrm{FP}}=0.1$
in \Fref{fig:Boot_N1000_M400}.

\begin{figure}
    \begin{minipage}{0.495\hsize}
    \centering
    \includegraphics[width=3in]{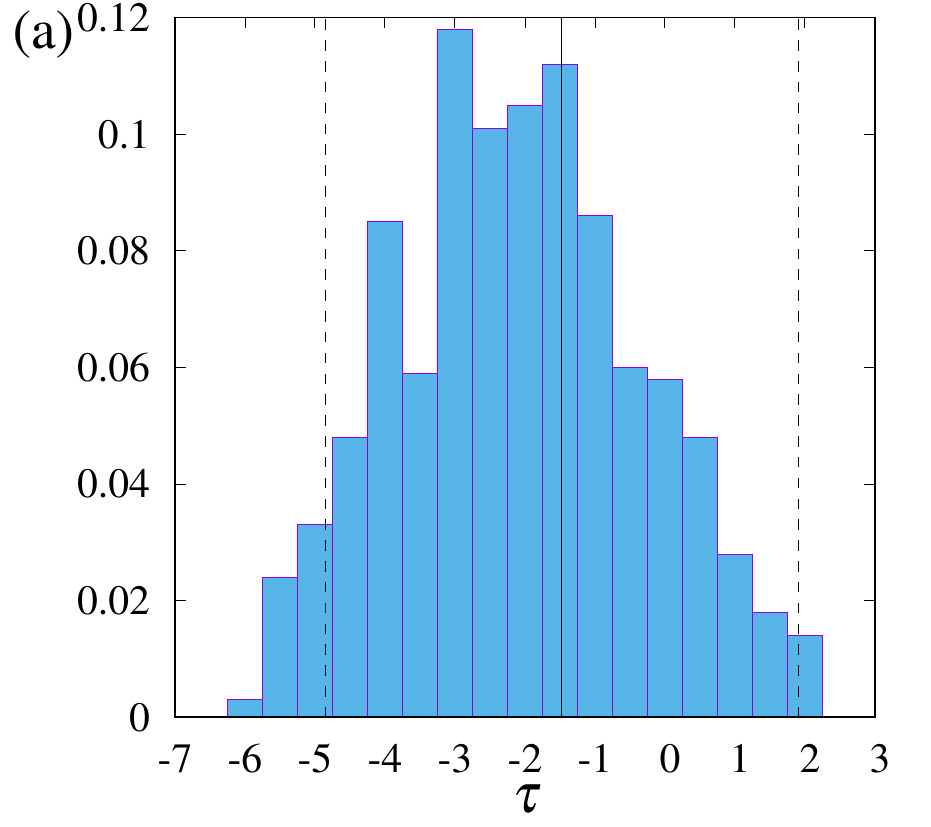}
    \end{minipage}
    \begin{minipage}{0.495\hsize}
    \centering
    \includegraphics[width=3in]{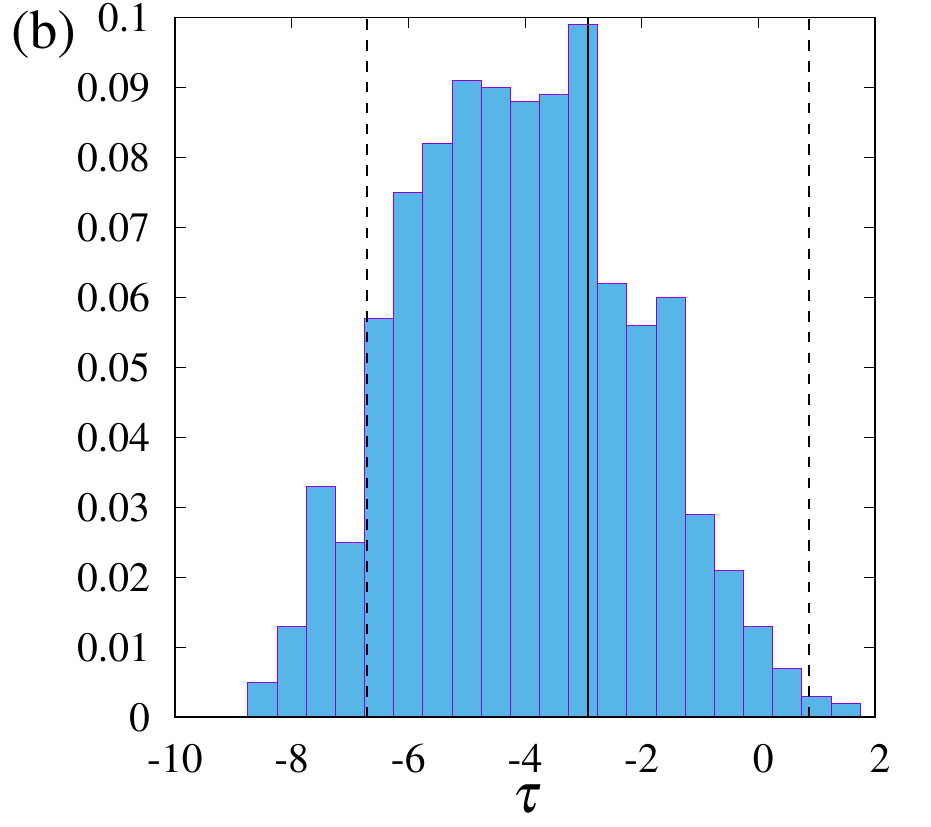}
    \end{minipage}
    \caption{Examples of bootstrap distribution at $N=1000$, $M=500$,
    $N_G=10$, $p_{\mathrm{TP}}=0.95$, and $p_{\mathrm{FP}}=0.1$. 
    The width of the bin is set at 0.5.
    The solid line and dashed lines represent
    $\hat{\tau}$ and the confidence interval, respectively.
    (a) Bootstrap distribution of $\tau$ for an infected patient
    where $\hat{X}_i^{\mathrm{(MAP)}}=0$ and $\hat{X}_i^{\mathrm{(Boot)}}=1$.
    (b) Bootstrap distribution of $\tau$ for a non-infected patient
    where $\hat{X}_i^{\mathrm{(MAP)}}=0$ and $\hat{X}_i^{\mathrm{(Boot)}}=1$.}
    \label{fig:Boot_hist}
\end{figure}

For the intuitive understanding of the bootstrap estimator,
I show examples of the bootstrap distributions of $\tau$
in \Fref{fig:Boot_hist} at $N=1000$, $M=500$, $N_G=10$, $p_{\mathrm{TP}}=0.95$, and $p_{\mathrm{FP}}=0.1$,
where the solid line represents $\hat{\tau}$ and the
two dashed lines indicate the confidence interval.
This histogram was obtained from 1000 bootstrap samples; 
note that the confidence interval \eref{eq:conf_int} is not that for the 
bootstrap distribution.
\Fref{fig:Boot_hist}(a) shows the
bootstrap distribution of an infected patient who is 
judged as non-infected by the MAP estimator and 
as infected by the bootstrap estimator.
\Fref{fig:Boot_hist}(b) shows the same for a
non-infected patient.
The patients shown in \Fref{fig:Boot_hist} (a) and (b)
contribute to the increase of the TP and FP of the bootstrap estimate,
respectively.

Another construction method of the credible interval 
is the use of the bootstrap percentile as
\begin{align}
\tau\in[G_B^{-1}(0.025),G_B^{-1}(0.975)],
\end{align}
where $G_B^{-1}(\alpha)$ is the $\alpha$-percentile of the bootstrap distribution.
I tried this interval for determining 
the infected patients.
The obtained TP is comparable with the normal theory;
however, the FP tends to be large compared with the interval determined by the normal theory.

\section{Estimation of unknown parameters by expectation maximization}
\label{sec:EM}

In this section, we consider the estimation of the unknown parameters:
prevalence $\rho$, TP probability $p_{\mathrm{TP}}$, and 
FP probability $p_{\mathrm{FP}}$. 
We construct their estimator by the maximum likelihood method,
where the likelihood is given by
\begin{align}
    \sum_{\bm{X}}P(\bm{Y}|\bm{X},p_{\mathrm{TP}},p_{\mathrm{FP}})P(\bm{X}|\rho)=P(\bm{Y}|\rho,p_{\mathrm{TP}},p_{\mathrm{FP}}),
\end{align}
and the estimators are given by
\begin{align}
    \hat{\rho}&=\arg\max_{\rho} \ln P(\bm{Y}|\rho,p_{\mathrm{TP}},p_{\mathrm{FP}})\\
    \hat{p}_{\mathrm{TP}}&=\arg\max_{p_{\mathrm{TP}}}\ln P(\bm{Y}|\rho,p_{\mathrm{TP}},p_{\mathrm{FP}})\\
    \hat{p}_{\mathrm{FP}}&=\arg\max_{p_{\mathrm{FP}}}\ln P(\bm{Y}|\rho,p_{\mathrm{TP}},p_{\mathrm{FP}}).
\end{align}
An approximation of the log-likelihood is given by the BP algorithm as 
Bethe free entropy \cite{Mezard-Montanari},
defined as
\begin{align}
    {\cal S}=\sum_{\mu=1}^M\ln{\cal Z}_\mu+\sum_{i=1}^N\ln{\cal Z}_i-\sum_{\mu}\sum_{i\in{\cal M}(\mu)}\ln{\cal Z}_{\mu i},
\end{align}
where
\begin{align}
\nonumber
    {\cal Z}_\mu&\equiv\sum_{\bm{X}}\prod_{i\in{\cal M}(\mu)} m_{i\to\mu}(X_i)P(Y_\mu|\bm{X})\\
    &=U_\mu\left(1-\tilde{q}_\mu\right)+W_\mu\tilde{q}_\mu\\
    \nonumber
    {\cal Z}_i&\equiv\sum_{X_i}\prod_{\mu\in{\cal G}(i)}\tilde{m}_{\mu\to i}(X_i)\{\rho X_i+(1-\rho)(1-X_i)\}\\
    &=\rho\prod_{\mu\in{\cal G}(i)}\tilde{\theta}_{\mu\to i}+(1-\rho)\prod_{\mu\in{\cal G}(i)}(1-\tilde{\theta}_{\mu\to i})\\
    \nonumber
    {\cal Z}_{\mu i}&\equiv \sum_{X_i} m_{i\to\mu}(X_i)\tilde{m}_{\mu\to i}(X_i)\\
    &=\theta_{i\to\mu}\tilde{\theta}_{\mu\to i}+(1-\theta_{i\to\mu})(1-\tilde{\theta}_{\mu\to i}),
\end{align}
and 
\begin{align}
    \tilde{q}_\mu=\prod_{i\in{\cal M}(\mu)}(1-\theta_{i\to\mu}).
\end{align}
We derive the maximum-likelihood estimator 
by the stationary condition of the Bethe free entropy \cite{Krzakala2012}.
After the calculation shown in Appendix\ref{sec:EM_estimator},
we obtain 
\begin{align}
    \hat{\rho}&=\frac{1}{N}\sum_{i=1}^N\hat{\theta}_i\label{eq:rhoh}\\
    \hat{p}_{\mathrm{TP}}&=\frac{\frac{1}{M}\sum_{\mu=1}^M\langle \mathbb{I}(Y_\mu=1,T_\mu(\bm{X}_{(\mu)})=1)\rangle_\mu}
    {\frac{1}{M}\sum_{\mu=1}^M\langle\mathbb{I}(T_\mu(\bm{X}_{(\mu)})=1)\rangle_\mu}\label{eq:u_hat}\\
    \hat{p}_{\mathrm{FP}}&=\frac{\frac{1}{M}\sum_{\mu=1}^M\langle\mathbb{I}(Y_\mu=1,T_\mu(\bm{X}_{(\mu)})=0)\rangle_\mu}
    {\frac{1}{M}\sum_{\mu=1}^M\langle\mathbb{I}(T_\mu(\bm{X}_{(\mu)})=0)\rangle_\mu},\label{eq:w_hat}
\end{align}
where $\langle\cdot\rangle_\mu$ denotes the expectation of 
$\bm{X}_{(\mu)}\equiv\{X_i|i\in {\cal M}(\mu)\}$
according to the posterior distribution with respect to the $\mu$-th test
defined by
\begin{align}
    P_\mu(\bm{X}_{(\mu)}|Y_\mu)=\frac{1}{{\cal Z}_\mu}P(Y_\mu|\bm{X}_{(\mu)})\prod_{i\in{\cal M}(\mu)} m_{i\to\mu}(X_i),
\end{align}
and 
\begin{align}
    \langle \mathbb{I}(Y_\mu=1,T_\mu(\bm{X}_{(\mu)})=1)\rangle_\mu&=
\frac{p_{\mathrm{TP}}Y_\mu\left(1-\tilde{q}_\mu\right)}{{\cal Z}_\mu}\\
    \langle \mathbb{I}(Y_\mu=1,T_\mu(\bm{X}_{(\mu)})=0)\rangle_\mu&=
    \frac{p_{\mathrm{FP}}Y_\mu\tilde{q}_\mu}{{\cal Z}_\mu}\\
    \langle\mathbb{I}(T_\mu(\bm{X}_{(\mu)})=1)\rangle_\mu&=\frac{U_\mu\left(1-\tilde{q}_\mu\right)}{{\cal Z}_\mu}\\
    \langle\mathbb{I}(T_\mu(\bm{X}_{(\mu)})=0)\rangle_\mu&=\frac{W_\mu\tilde{q}_\mu}{{\cal Z}_\mu}.
\end{align}
Eqs. (\ref{eq:u_hat})--(\ref{eq:w_hat}) always 
have trivial fixed points at $0$ and $1$,
and to avoid these solutions,
we solve following expressions:
\begin{align}
f(u,v)&\equiv\sum_\mu\frac{(2Y_\mu-1)\left(1-\tilde{q}_\mu\right)}{{\cal Z}_\mu}=0
\label{eq:f_uv}\\
g(u,v)&\equiv\sum_\mu\frac{(2Y_\mu-1)\tilde{q}_\mu}{{\cal Z}_\mu} = 0,\label{eq:g_uv}
\end{align}
which are equivalent to eqs. (\ref{eq:u_hat})--(\ref{eq:w_hat}).
Note that the extremization conditions of eqs.(\ref{eq:rhoh})--(\ref{eq:w_hat})
correspond to the Nishimori line \cite{Iba1999, Nishimori}.

We calculate the infected probability of patients and 
the estimators of the unknown parameters by the
EM method.
In the E-step, the fixed points of the BP,
$\{\tilde{\theta}_{\mu\to i}\}$ and 
$\{\theta_{i\to\mu}\}$, are achieved by recursive updating
under a fixed $\hat{p}_{\mathrm{TP}},~\hat{p}_{\mathrm{FP}}$ and $\hat{\rho}$.
In the M-step,
these parameters are updated according to
the extremization conditions of eqs. (\ref{eq:rhoh})--(\ref{eq:w_hat}).
We term this method the BP+EM algorithm, which is 
summarized in Algorithm \ref{alg:BP_with_EM}.
We solve \eref{eq:f_uv} and \eref{eq:g_uv} using the Newton method in the M-step;
however, the optimization at each M-step induces algorithmic instability.
Hence, we update $\hat{p}_{\mathrm{TP}}$ and $\hat{p}_{\mathrm{FP}}$ 
for only one step following the Newton method.

\begin{figure}
    \begin{minipage}{0.325\hsize}
    \centering
    \includegraphics[width=2.1in]{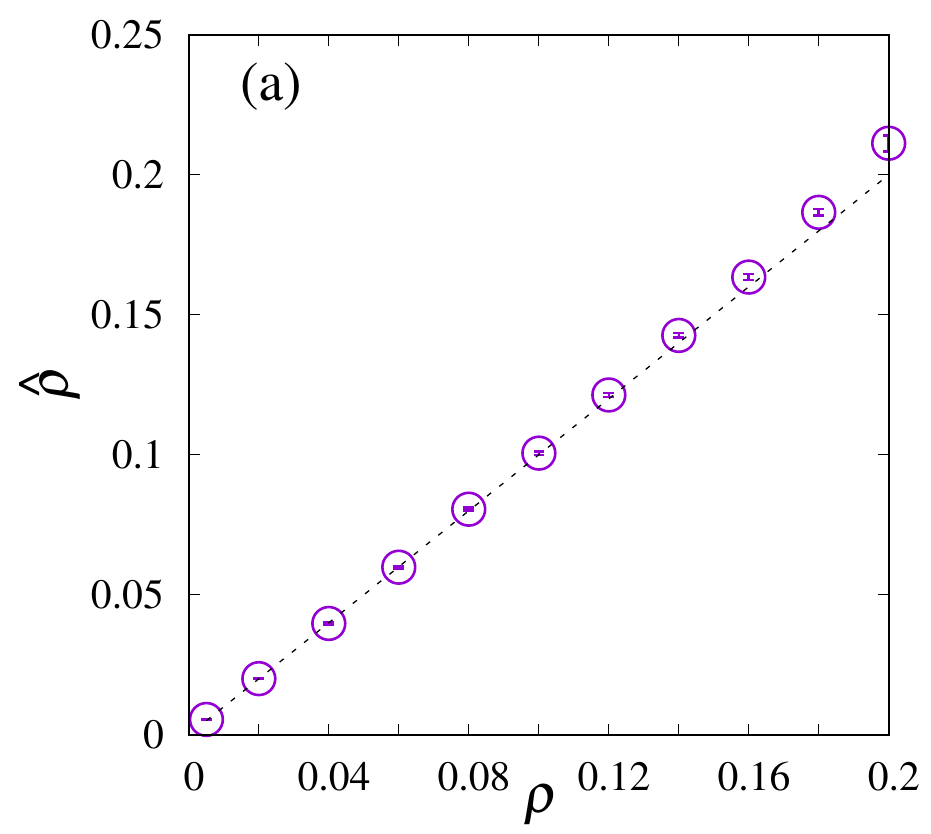}
    \end{minipage}
    \begin{minipage}{0.325\hsize}
    \centering
    \includegraphics[width=2.1in]{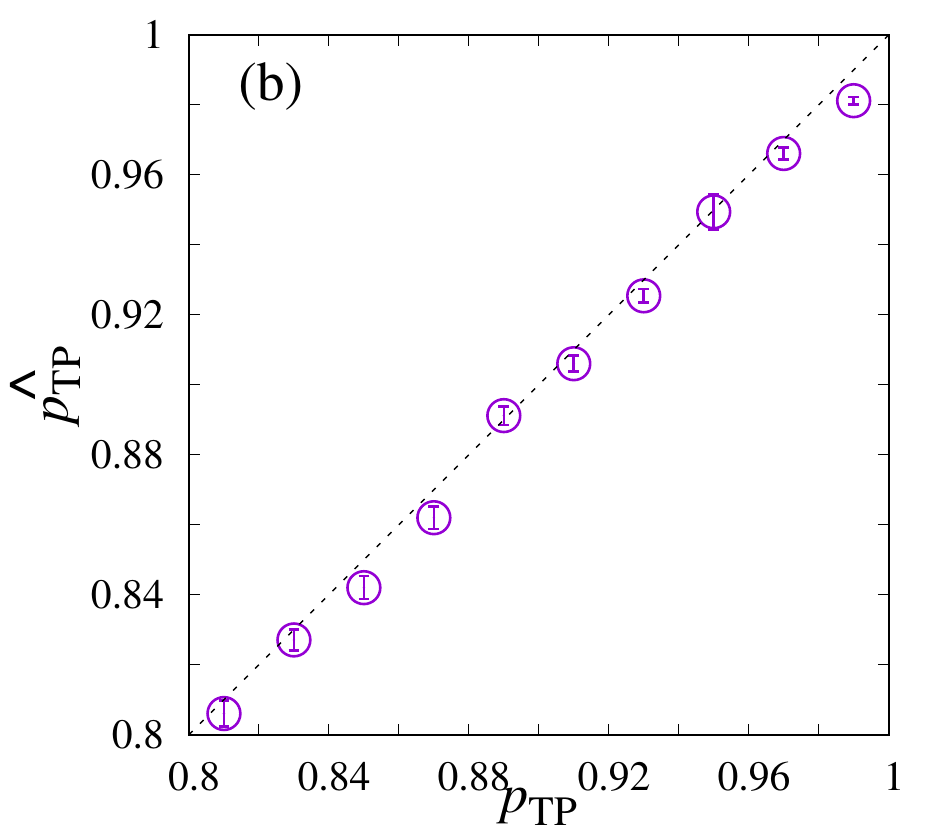}
    \end{minipage}
    \begin{minipage}{0.325\hsize}
    \centering
      \includegraphics[width=2.1in]{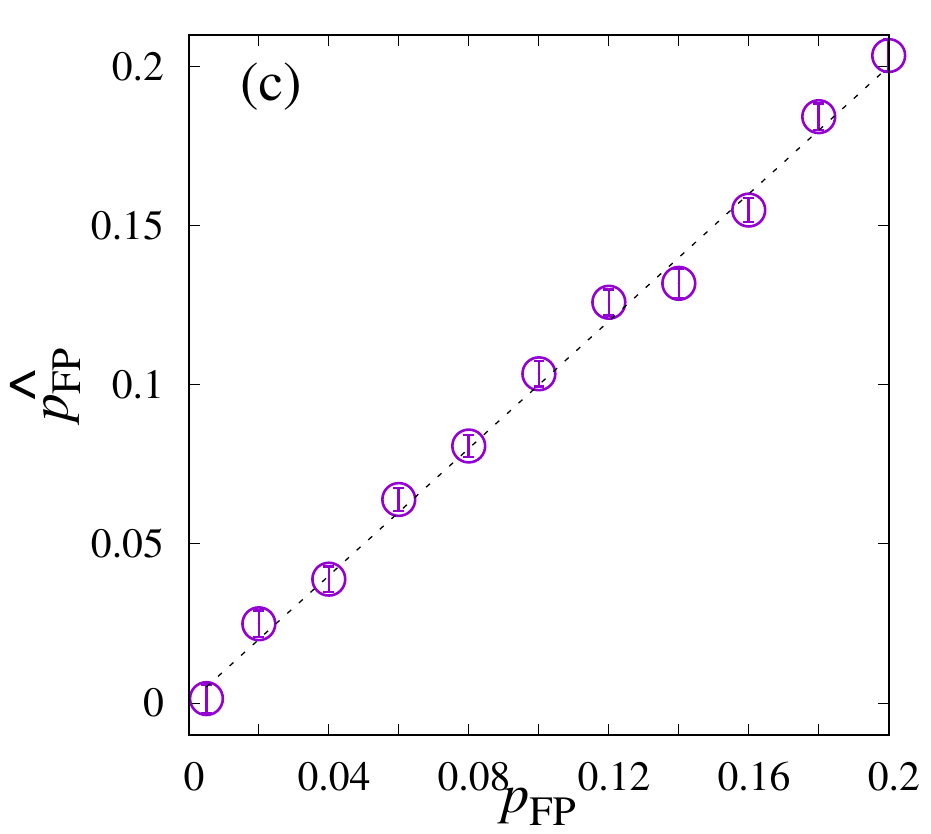}
      \end{minipage}
    \caption{Comparison between true values and estimated values of hyperparameters.
    The gradient of the diagonal lines is 1, and 
    system sizes are $N=1000,~M=500$, and $N_G=10$.
    (a) $\hat{\rho}$ vs. $\rho$ plot at $p_{\mathrm{TP}}=0.95$ and $p_{\mathrm{FP}}=0.1$. The parameters $p_{\mathrm{TP}}$ and $p_{\mathrm{FP}}$ are estimated by the EM method.
    (b) $\hat{p}_{\mathrm{TP}}$ vs. $p_{\mathrm{TP}}$ plot at $\rho=0.1$, and $p_{\mathrm{FP}}=0.1$, where
     $\rho$ and $p_{\mathrm{FP}}$ are estimated by the EM method.
    (c) $\hat{p}_{\mathrm{FP}}$ vs. $p_{\mathrm{FP}}$ plot at $\rho=0.1$ and $p_{\mathrm{TP}}=0.95$,
    where $\rho$ and $p_{\mathrm{TP}}$ are estimated by the EM method.
    }
    \label{fig:hypara}
\end{figure}

\Fref{fig:hypara} show the comparison between 
estimated parameters and true parameters at
$N=1000$, $M=500$, and $N_G=10$
for (a) $\rho$ at $p_{\mathrm{TP}}=0.95$ and $p_{\mathrm{FP}}=0.1$,
(b) $p_{\mathrm{TP}}$ at $\rho=0.1$ and $p_{\mathrm{FP}}=0.1$,
and (c) $p_{\mathrm{FP}}$ at $\rho=0.1$ and $p_{\mathrm{TP}}=0.95$.
The gradient of the diagonal lines is 1.
Hence, the point on this line indicates that
the accurate estimation of the unknown parameters is achieved.
In all the figures, parameters that are not shown in the figures
are also estimated simultaneously.
For the entire parameter region, the M-step converges to the true parameter.
The TP rate and FP rate of the BP+EM algorithm
are the same as those of the BP algorithm where the parameters are known.

I note that the behavior of the BP+EM algorithm
heavily depends on the initial condition of $\hat{p}_{\mathrm{TP}}$ and $\hat{p}_{\mathrm{FP}}$.
When the initial conditions of 
$\hat{p}_{\mathrm{TP}}$ and $\hat{p}_{\mathrm{FP}}$ are close to their true values,
the BP+EM algorithm is stable;
hence, the proposed method should be treated as a correction 
of the experimentally estimated values.
The estimation of $\hat{\rho}$ is insensitive to the initial condition,
which is smaller than $\alpha$.

\begin{algorithm}[H]
\caption{BP+EM for Bayesian group testing}
\label{alg:BP_with_EM}
\begin{algorithmic}[1]
\Require {$\bm{Y}\sim P(\bm{Y}|\bm{X}^{(0)})$ and $\bm{F}$}
\Ensure {$\bm{\theta}\in[0,1]^N$}
      \Initialize{
      \{$\theta^{(0)}_{i\to\mu}\}\gets$ initial value from $[0,1]^{N\times M}$ \\
      $\{\tilde{\theta}^{(0)}_{\mu\to i}\}\gets$ initial value from $[0,1]^{M\times N}$ \\
      $\{\hat{\rho}^{[0]},\hat{p}_{\mathrm{TP}}^{[0]},\hat{p}_{\mathrm{FP}}^{[0]}\}\gets$ initial value from $[0,1]^3$ \\
      $\bm{U}^{[0]}\gets\hat{p}_{\mathrm{TP}}^{[0]}\bm{Y}+(1-\hat{p}_{\mathrm{TP}}^{[0]})(\bm{1}_M-\bm{Y}),~
      \bm{W}^{[0]}\gets\hat{p}_{\mathrm{FP}}^{[0]}\bm{Y}+(1-\hat{p}_{\mathrm{FP}}^{[0]})(\bm{1}_M-\bm{Y})$ 
      }
      \For{$s = 1 \, \ldots \, S$}
        \For{$t = 1 \, \ldots \, T$}
            \For {all combinations of $(\mu,i)$ such that $F_{\mu i}=1$}
                \State{$\tilde{Z}_{\mu\to i}^{(t)}\gets U^{[s-1]}_\mu\left\{2-\prod_{j\in{\cal M}(\mu)\backslash i}\left(1-\theta_{j\to\mu}^{(t-1)}\right)\right\}
                +W^{[s-1]}_\mu\prod_{j\in{\cal M}(\mu)\backslash i}\left(1-\theta_{j\to\mu}^{(t-1)}\right)$}
                \State{$Z_{i\to\mu}^{(t)}\gets\hat{\rho}^{[s-1]} \prod_{\nu\in{\cal G}(i)\backslash \mu} \tilde{\theta}_{\nu\to i}^{(t-1)}+(1-\hat{\rho}^{[s-1]})\prod_{\nu\in{\cal G}(i)\backslash \mu} \left(1-\tilde{\theta}_{\nu\to i}^{(t-1)}\right)$}
                \State{$\tilde{\theta}_{\mu\to i}^{(t)}\gets\frac{U^{[s-1]}_\mu}{\tilde{Z}^{(t)}_{\mu\to i}}$}
                \State{$\theta_{i\to\mu}^{(t)}\gets\frac{\hat{\rho}^{[s-1]} \prod_{\nu\in{\cal G}(i)\backslash \mu} \tilde{\theta}_{\nu\to i}^{(t-1)}}{Z^{(t)}_{i\to\mu}}$}
            \EndFor
        \EndFor
            \For{$i = 1 \, \ldots \, N$}
                \State{$\hat{\theta}_i^{[s]}\gets$ $\frac{\hat{\rho}^{[s-1]}\prod_{\mu\in{\cal G}(i)}\tilde{\theta}_{\mu\to i}^{(T)}}
                {\hat{\rho}^{[s-1]}\prod_{\mu\in{\cal G}(i)}\tilde{\theta}_{\mu\to i}^{(T)}
                +(1-\hat{\rho}^{[s-1]})\prod_{\mu\in{\cal G}(i)}\left(1-\tilde{\theta}_{\mu\to i}^{(T)}\right)}$}
            \EndFor
            \State{$\hat{\rho}^{[s]}\gets\frac{1}{N}\sum_i\hat{\theta}_i^{[s]}$}
        \For{$\mu = 1 \, \ldots \, M$}
            \State{$\tilde{q}_\mu^{[s]}\gets\prod_{i\in{\cal M}(\mu)}(1-\theta_{i\to\mu}^{(T)})$}
            \State{${\cal Z}^{[s]}_\mu\gets U_\mu^{[s-1]}(1-\tilde{q}_\mu^{[s]})+W_\mu^{[s-1]}\tilde{q}_\mu^{[s]}$}
        \EndFor
        \State{$f^{[s]}\gets\sum_\mu\frac{(Y_\mu-(1-Y_\mu))(1-\tilde{q}_\mu^{[s]})}{{\cal Z}_\mu^{[s]}},
        ~g^{[s]}\gets\sum_\mu\frac{(Y_\mu-(1-Y_\mu))\tilde{q}_\mu^{[s]}}{{\cal Z}_\mu^{[s]}}$}
        \State{$G^{[s]}\gets -
        \begin{bmatrix}
        \sum_\mu\frac{(2Y_\mu-1)^2\left(1-\tilde{q}^{[s]}_\mu\right)^2}{{{\cal Z}^{[s]}_\mu}^2} & 
        \sum_\mu\frac{(2Y_\mu-1)^2\left(1-\tilde{q}^{[s]}_\mu\right)\tilde{q}^{[s]}_\mu}{{{\cal Z}^{[s]}_\mu}^2} \\
        \sum_\mu\frac{(2Y_\mu-1)^2\left(1-\tilde{q}^{[s]}_\mu\right)\tilde{q}^{[s]}_\mu}{{{\cal Z}^{[s]}_\mu}^2} & 
        \sum_\mu\frac{(2Y_\mu-1)^2(\tilde{q}_\mu^{[s]})^2}
        {{{\cal Z}_\mu^{[s]}}^2} \\
        \end{bmatrix}
        $}
        \State{$[\hat{p}_{\mathrm{TP}}^{[s]},\hat{p}_{\mathrm{FP}}^{[s]}]^{\mathrm{T}}
        \gets [\hat{p}_{\mathrm{TP}}^{[s-1]},\hat{p}_{\mathrm{FP}}^{[s-1]}]^{\mathrm{T}}-{G^{[s]}}^{-1}[f^{[s]},g^{[s]}]^{\mathrm{T}}$}
        \State{$\bm{U}^{[s]}\gets~\hat{p}_{\mathrm{TP}}^{[s]}\bm{Y}+(1-\hat{p}_{\mathrm{TP}}^{[s]})(\bm{1}_M-\bm{Y}),~
        \bm{W}^{[s]}\gets~\hat{p}_{\mathrm{FP}}^{[s]}\bm{Y}+(1-\hat{p}_{\mathrm{FP}}^{[s]})(\bm{1}_M-\bm{Y})$}
    \EndFor
\end{algorithmic}
\end{algorithm}

\section{Hierarchical Bayes approach}
\label{sec:HBP}

\begin{figure}
    \centering
    \includegraphics[width=4in]{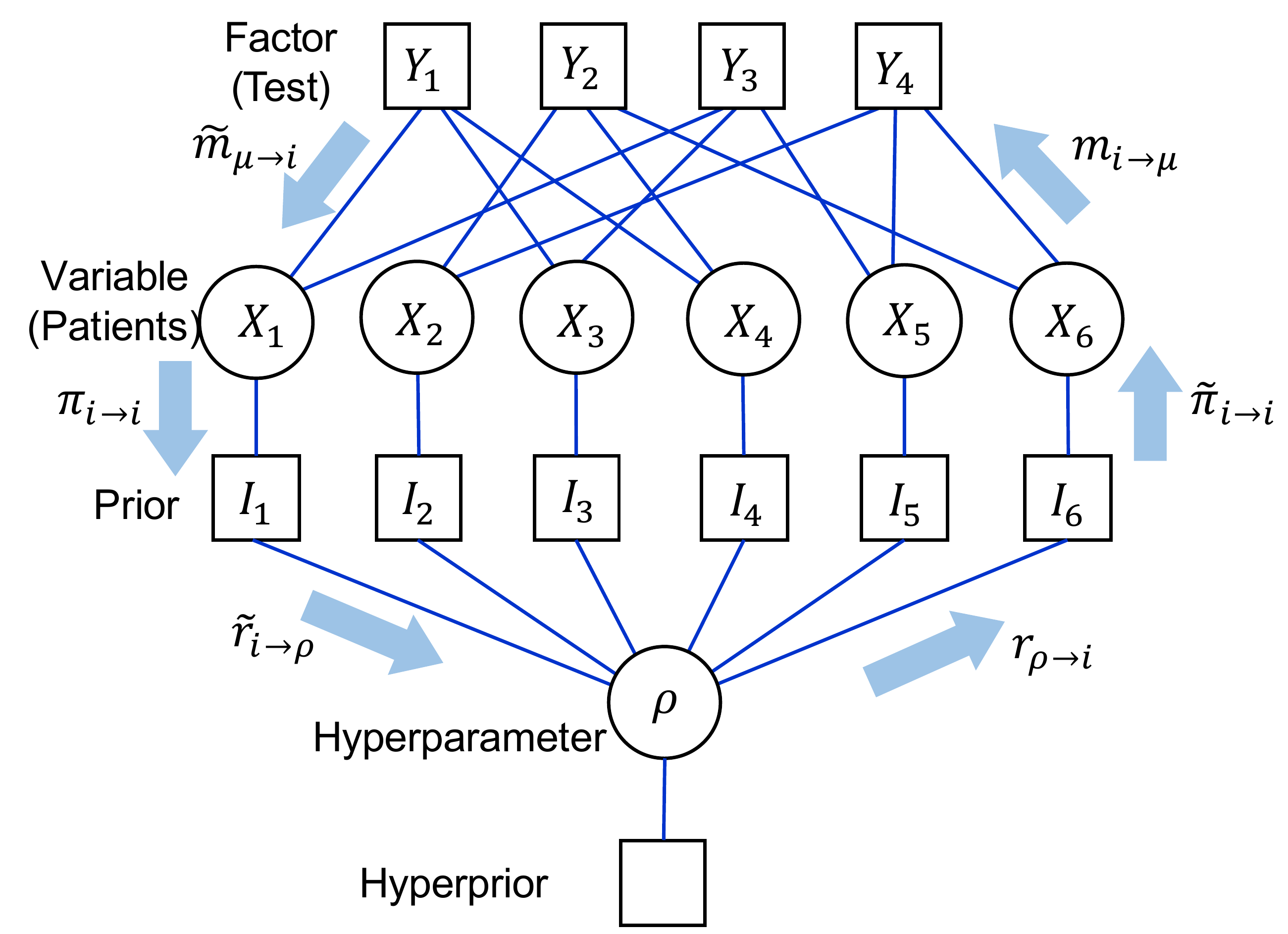}
    \caption{Graphical representation and messages of the hierarchical Bayes model
    for the group testing at $N=6$, $M=4$, $N_G=3$, and $N_O=2$.}
    \label{fig:Hierarchical_Bayes}
\end{figure}

As another approach to estimating prevalence,
we introduce the hierarchical Bayes model,
where the prevalence is regarded as a hyperparameter 
distributed according to the hyperprior distribution
\begin{align}
    \phi(\rho;a,b)=\frac{\rho^{a-1}(1-\rho)^{b-1}}{B(a,b)},
\end{align}
which is the beta distribution with the hyperhyperparameters $a$ and $b$,
and $B(a,b)$ is the beta function.
The beta distribution is the conjugate of the Bernoulli distribution.
A graphical representation of group testing for the hierarchical Bayes model is
shown in \Fref{fig:Hierarchical_Bayes}.
The prior distribution of $X_i$ under a given $\rho$
is regarded as an ``interaction''
that is represented by a factor node $I_i$.
We introduce additional messages $\pi_{i\to i}$, $\tilde{\pi}_{i\to i}$,
$\tilde{r}_{i\to\rho}$, and $r_{\rho\to i}$ for all $i$,
that are propagated from $X_i$ to $I_i$, $I_i$ to $X_i$, 
$I_i$ to $\rho$, and $\rho$ to $I_i$, respectively, as shown in \Fref{fig:Hierarchical_Bayes}.

The messages propagated between the bipartite graph of $\bm{Y}$ and $\bm{X}$
are given by
\begin{align}
    \tilde{m}_{\mu\to i}(X_i)&\propto\sum_{\bm{X}\backslash X_i}P(Y_\mu|\bm{X})\prod_{j\in{\cal M}(\mu)\backslash i}m_{j\to\mu}(X_i)\\
    m_{i\to\mu}(X_i)&=\tilde{\pi}_{i\to i}(x_i)\prod_{\nu\in{\cal G}(i)\backslash \mu}\tilde{m}_{\nu\to i}(x_i),
\end{align}
where $\tilde{\pi}_{i\to i}$ carries the prior information to $X_i$. 
Here, we express $\tilde{\pi}_{i\to i}(x_i)$ by one parameter $\tilde{\rho}_i$, which is derived later, as
\begin{align}
    \tilde{\pi}_{i\to i}=\tilde{\rho}_iX_i+(1-\tilde{\rho}_i)(1-X_i).\label{eq:vh}
\end{align}
Using \eref{eq:vh}, the parameters $\tilde{\theta}_{\mu\to i}$ and 
$\theta_{i\to\mu}$ that express the messages as eqs. (\ref{eq:one_para_h})--(\ref{eq:one_para}),
are given by
\begin{align}
    \tilde{\theta}_{\mu\to i}&=\frac{U_\mu}{\tilde{Z}_{\mu\to i}}\label{eq:thetah_HGT}\\
    \theta_{i\to\mu}&=\frac{\tilde{\rho}_i\prod_{\nu\in{\cal G}(i)\backslash \mu} \tilde{\theta}_{\nu\to i}}{Z_{i\to\mu}},\label{eq:theta_HGT}
\end{align}
where
\begin{align}
    \tilde{Z}_{\mu\to i}&=U_\mu\left(2-\prod_{j\in{\cal M}(\mu)\backslash i} (1-\theta_{j\to\mu})\right)
    +W_\mu\prod_{j\in{\cal M}(\mu)\backslash i}(1-\theta_{j\to\mu})\\
    Z_{i\to\mu}&=\tilde{\rho}_i\prod_{\nu\in{\cal G}(i)\backslash \mu} \tilde{\theta}_{\nu\to i}+(1-\tilde{\rho}_i)\prod_{\nu\in{\cal G}(i)\backslash \mu} (1-\tilde{\theta}_{\nu\to i}),
\end{align}
and $U_\mu$ and $W_\mu$ are given by eqs.(\ref{eq:U})--(\ref{eq:W}).
The messages between the variables and priors are given by
\begin{align}
\pi_{i\to i}(X_i)&\propto\prod_{\mu\in G(i)}\tilde{m}_{\mu\to i}(x_i)\\
    \tilde{\pi}_{i\to i}(x_i)&=\int_0^1 d\rho\{\rho X_i+(1-\rho)(1-X_i)\}r_{\rho\to i}(\rho),
\end{align}
and we obtain
\begin{align}
    \tilde{\rho}_i=\int_0^1 d\rho~\rho r_{\rho\to i}(\rho).
    \label{eq:xih}
\end{align}
Further, by setting
\begin{align}
    \pi_{i\to i}(X_i)=\pi_i X_i+(1-\pi_i)(1-X_i),
    \label{eq:v}
\end{align}
we obtain
\begin{align}
    {\pi}_i=\frac{\prod_{\mu\in{\cal G}(i)}\tilde{\theta}_{\mu\to i}}
    {\prod_{\mu\in{\cal G}(i)}\tilde{\theta}_{\mu\to i}+\prod_{\mu\in{\cal G}(i)}(1-\tilde{\theta}_{\mu\to i})},
    \label{eq:xi}
\end{align}
which corresponds to the infection probability when the prior is ignored.
The messages between prior $I_i$ and the hyperparameter $\rho$ are given by
\begin{align}
\nonumber
    \tilde{r}_{i\to\rho}(\rho)&\propto\sum_{X_i}\{\rho X_i+(1-\rho)(1-X_i)\}\pi_{i\to i}(X_i)\\
    &=\rho\pi_i+(1-\rho)(1-\pi_i)\\
    r_{\rho\to i}(\rho)&\propto\phi(\rho)\prod_{j\neq i}\tilde{r}_{j\to\rho}(\rho).\label{eq:s}
\end{align}
Using these messages,
we can approximate the marginal distribution as
\begin{align}
    P(X_i)\propto \tilde{\pi}_{i\to i}(X_i)\prod_{\mu\in{\cal G}(i)}\tilde{m}_{\mu\to i}(X_i),
\end{align}
and the infection probability of $X_i$ is estimated as
\begin{align}
    \hat{\theta}_i=\frac{\tilde{\rho}_i\prod_{\mu\in{\cal G}(i)}\tilde{\theta}_{\mu\to i}}
    {\tilde{\rho}_i\prod_{\mu\in{\cal G}(i)}\tilde{\theta}_{\mu\to i}
    +(1-\tilde{\rho}_i)\prod_{\mu\in{\cal G}(i)}\left(1-\tilde{\theta}_{\mu\to i}\right)}.
    \label{eq:theta}
\end{align}
We refer to the BP algorithm for the hierarchical Bayes model as the 
hierarchical BP (HBP) algorithm; its psuedocode is
summarized in Algorithm \ref{alg:BP_hierarchical}.
For HBP, we introduce an additional damping as
\begin{align}
\pi_i^{(t)}&\gets d\pi_i^{(t)}+(1-d)\pi_i^{(t-1)}\\
\tilde{\rho}_i^{(t)}&\gets d\tilde{\rho}_i^{(t)}+(1-d)\tilde{\rho}_i^{(t-1)}.
\end{align}

When the infection probability
of each patient can be guessed 
from symptoms before performing the test and 
has different values for each patient,
we can apply the information into the prior as $\rho_i$.
The BP algorithm for this case 
is obtained from the HBP by fixing the value
of $\tilde{\rho}_i$ at $\rho_i$.

\begin{algorithm}[H]
\caption{HBP for group testing}
\label{alg:BP_hierarchical}
\begin{algorithmic}[1]
\Require {$\bm{Y}\sim P(\bm{Y}|\bm{X}^{(0)})$ and $\bm{F}$}
\Ensure {$\bm{\theta}\in[0,1]^N$}
      \Initialize{
      \{$\theta^{(0)}_{i\to\mu}\}\gets$ initial value from $[0,1]^{N\times M}$ \\
      $\{\tilde{\theta}^{(0)}_{\mu\to i}\}\gets$ initial value from $[0,1]^{M\times N}$ \\
      $\bm{\pi}^{(0)}\gets$ initial value from $[0,1]^N$ \\
      $\tilde{\bm{\rho}}^{(0)}\gets$ initial value from $[0,1]^N$ 
      }
    \State{$\bm{U}\gets~u\bm{Y}+(1-u)(\bm{1}_M-\bm{Y})$,~~
    $\bm{W}\gets~w\bm{Y}+(1-w)(\bm{1}_M-\bm{Y})$}
    \For{$t = 1 \, \ldots \, T$}
        \For {all combinations of $(\mu,i)$ such that $F_{\mu i}=1$}
            \State{$\tilde{Z}_{\mu\to i}^{(t)}\gets U_\mu\left\{2-\prod_{j\in{\cal M}(\mu)\backslash i}\left(1-\theta_{j\to\mu}^{(t-1)}\right)\right\}
    +W_\mu\prod_{j\in{\cal M}(\mu)\backslash i}\left(1-\theta_{j\to\mu}^{(t-1)}\right)$}
            \State{$Z_{i\to\mu}^{(t)}\gets \tilde{\rho}_i^{(t-1)} \prod_{\nu\in{\cal G}(i)\backslash \mu} \tilde{\theta}_{\nu\to i}^{(t-1)}+(1-\tilde{\rho}_i^{(t-1)})\prod_{\nu\in{\cal G}(i)\backslash \mu} \left(1-\tilde{\theta}_{\nu\to i}^{(t-1)}\right)$}
            \State{$\tilde{\theta}_{\mu\to i}^{(t)}\gets\frac{U_\mu}{\tilde{Z}^{(t)}_{\mu\to i}}$}
            \State{$\theta_{i\to\mu}^{(t)}\gets\frac{\tilde{\rho}_i^{(t-1)} \prod_{\nu\in{\cal G}(i)\backslash \mu} \tilde{\theta}_{\nu\to i}^{(t-1)}}{Z^{(t)}_{i\to\mu}}$}
        \EndFor
        \For{$i=1 \, \ldots \, N$}
            \State{$\pi_i^{(t)}\gets\frac{\prod_{\mu\in{\cal G}(i)}\tilde{\theta}^{(t)}_{\mu\to i}}
    {\prod_{\mu\in{\cal G}(i)}\tilde{\theta}_{\mu\to i}^{(t)}+\prod_{\mu\in{\cal G}(i)}(1-\tilde{\theta}_{\mu\to i}^{(t)})}$}
            \State{$\Xi_{\rho\to i}^{(t)}\gets\int_0^1 d\rho \phi(\rho)\prod_{j\neq i}\{\rho\pi_j^{(t)}+(1-\rho)(1-\pi_j^{(t)})\}$}
            \State{$\tilde{\rho}^{(t)}_i\gets\frac{1}{\Xi_{\rho\to i}^{(t)}}\int_0^1 d\rho~\rho \phi(\rho)\prod_{j\neq i}\{\rho\pi_j^{(t)}+(1-\rho)(1-\pi_j^{(t)})\}$}
        \EndFor
    \EndFor
    \For{$i = 1 \, \ldots \, N$}
        \State{$\hat{\theta}_i\gets$ $\frac{\tilde{\rho}_i^{(T)}\prod_{\mu\in{\cal G}(i)}\tilde{\theta}_{\mu\to i}^{(T)}}
    {\tilde{\rho}_i^{(T)}\prod_{\mu\in{\cal G}(i)}\tilde{\theta}_{\mu\to i}^{(T)}
    +\left(1-\tilde{\rho}_i^{(T)}\right)\prod_{\mu\in{\cal G}(i)}\left(1-\tilde{\theta}_{\mu\to i}^{(T)}\right)}$}
    \EndFor
\end{algorithmic}
\end{algorithm}

\begin{figure}
    \begin{minipage}{0.495\hsize}
    \centering
    \includegraphics[width=2.7in]{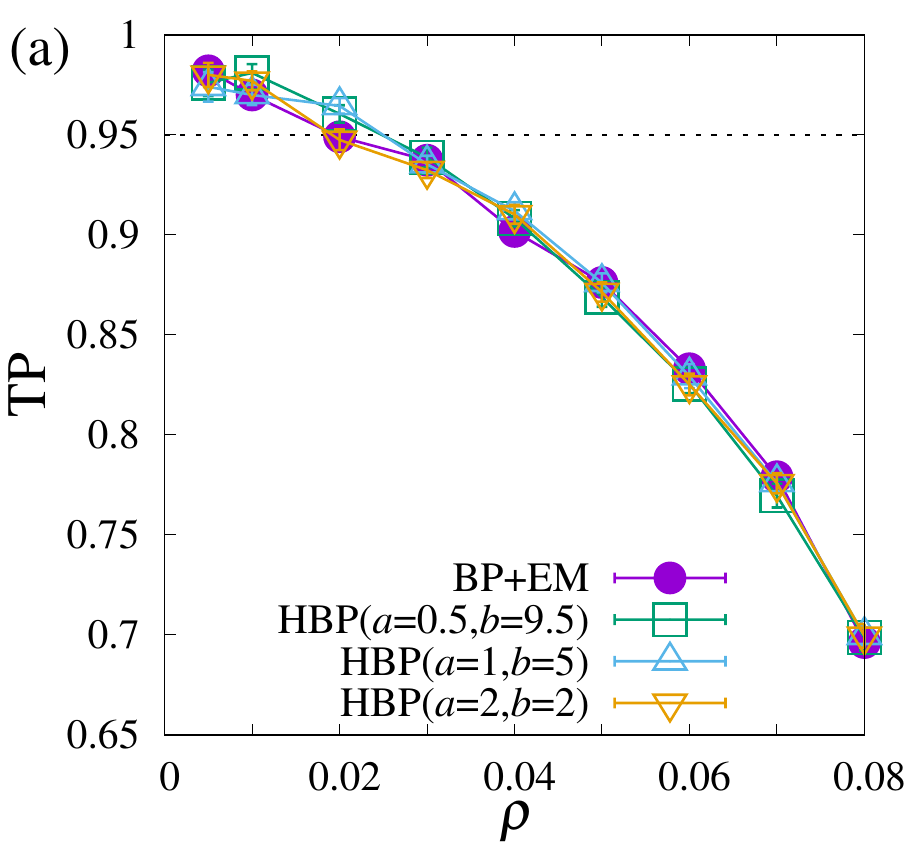}
    \end{minipage}
    \begin{minipage}{0.495\hsize}
    \centering
    \includegraphics[width=2.7in]{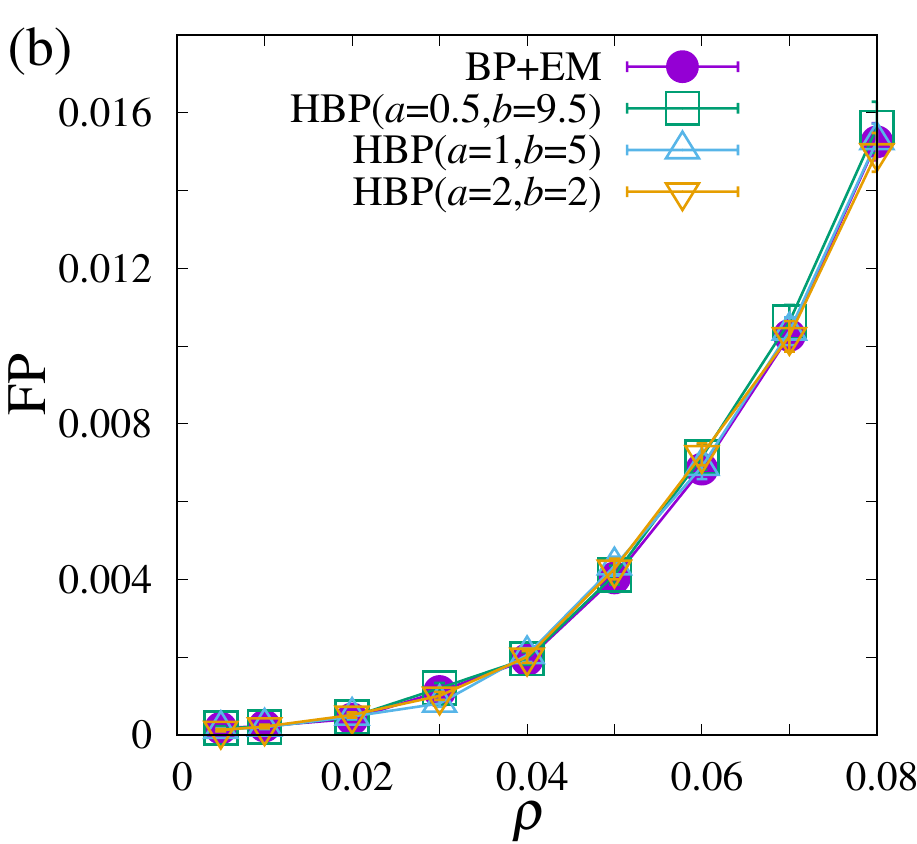}
    \end{minipage}
    \caption{Comparison of the BP+EM algorithm and HBP algorithm for (a) TP and (b) FP
    at $N=1000$, $M=500$, $N_G=10$, $p_{\mathrm{TP}}=0.95$, and $p_{\mathrm{FP}}=0.05$.
    The horizontal line in (a) indicates $\mathrm{TP}=p_{\mathrm{TP}}$.}
    \label{fig:HBP_TP_FP}
\end{figure}

\begin{figure}
\centering
    \includegraphics[width=3in]{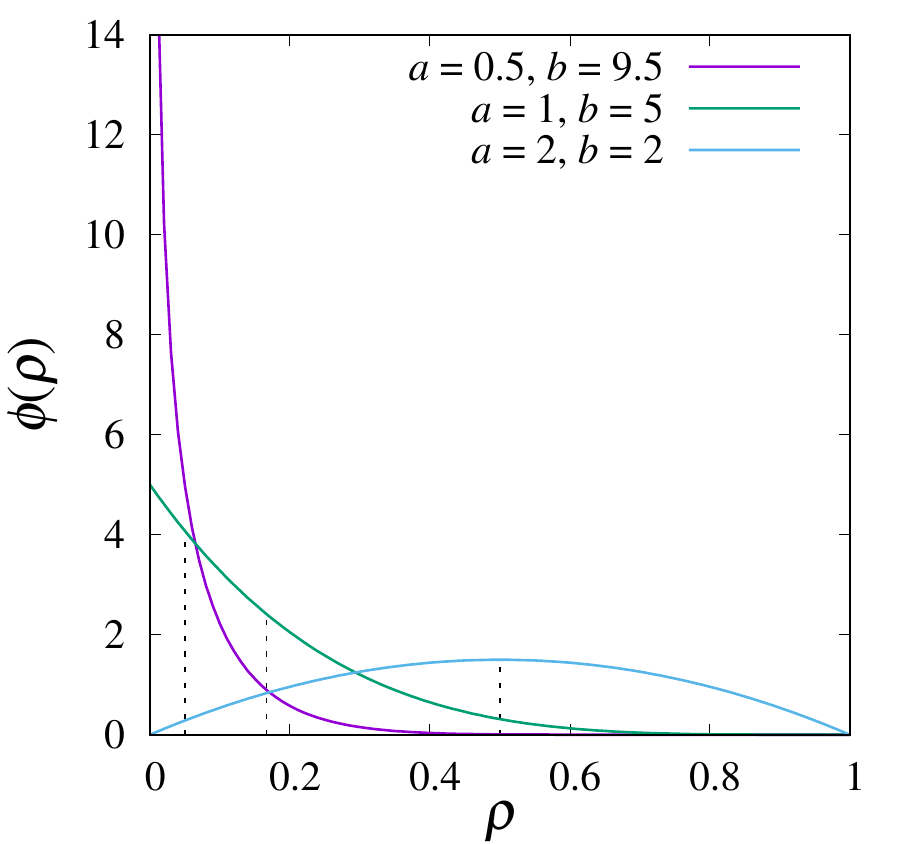}
    \caption{Beta distributions at $a=0.5,b=9.5$, 
    $a=1,b=5$, and $a=2,b=2$, which are used in \Fref{fig:HBP_TP_FP}
    as hyperprior. Their means are 0.05, $1\slash 6$, and $0.5$, respectively,
    denoted by dashed lines.}
    \label{fig:beta_dist}
\end{figure}

\Fref{fig:HBP_TP_FP} shows the 
comparison between the BP+EM algorithm and HBP algorithm
for the (a) TP and (b) FP obtained by the MAP estimator
at $N=1000$, $M=500$, $N_G=10$, $p_{\mathrm{TP}}=0.95$, and $p_{\mathrm{FP}}=0.05$.
Here, $p_{\mathrm{TP}}$ and $p_{\mathrm{FP}}$ are fixed at their true values.
As shown in this figure,
the TP and FP from the HBP algorithm have 
almost the same values as those from the BP+EM algorithm.
Further, the hyperhyperparameter in the beta distribution
has a small influence on the TP and FP.
The beta distributions used for \Fref{fig:HBP_TP_FP}
are shown in \Fref{fig:beta_dist}. 
The mean of the beta distribution is given by $a/(a+b)$; hence, 
the mean of the hyperprior for $a=0.5,b=0.95$, $a=1,b=5$, and $a=2,b=2$ are
0.05, $1\slash 6$, and $0.5$, respectively.
The $a=0.5, b=0.95$ case describes the 
high probability at small values of $\rho$.
The means of the other cases exceed the $\rho$-region
shown in \Fref{fig:HBP_TP_FP}.
In particular, the mode corresponds to 
the mean, 0.5, in $a=b=2$ case.
Despite such ``dense'' prior,
the reconstruction performance is comparable with the BP+EM method.

\subsection{Comparison of BP+EM and HBP for finite system size}

In this section, we discuss the difference between the BP+EM algorithm and HBP algorithm
as estimation methods for prevalence.
First,
we consider the $N\to\infty$ limit,
where the saddle point method can be applied to 
the integral of $\rho$ in \eref{eq:xih}.
After the calculation shown in Appendix\ref{sec:saddle},
we obtain
\begin{align}
\tilde{\rho}_i=\rho^*_i,
\end{align}
where $\rho^*_i$ satisfies
\begin{align}
\rho^*_i&=\frac{1}{N-1}\sum_{j\neq i}\tilde{\theta}_j(\rho_i^*),\label{eq:rhoh_HBP}\\
\tilde{\theta}_j(\rho)&=\frac{\rho\prod_{\mu\in{\cal G}(j)}\tilde{\theta}_{\mu\to j}}{\rho\prod_{j\neq i}\tilde{\theta}_{\mu\to j}+(1-\rho)\prod_{j\neq i}(1-\tilde{\theta}_{\mu\to j})}.\label{eq:theta_tilde}
\end{align}
We note that \eref{eq:rhoh_HBP} does not depend on the hyperprior
and the prevalence is estimated as $\hat{\rho}=\frac{1}{N}\sum_{i=1}^{N}\tilde{\rho}_i$.
Comparing the estimated prevalence in the HBP algorithm with that of the
BP+EM algorithm, \eref{eq:rhoh} shows that
the difference between the two estimators is negligible at $N\to\infty$.
Therefore, we compare the BP+EM and HBP algorithms by
focusing on the following aspects.

\begin{description}

\item{(I) } Accuracy as an estimator of the prevalence for finite $N$

As mentioned previously, 
the difference between the two estimators is negligible at $N\to\infty$.
However, the two estimators do not coincide with each other at finite $N$.
We quantify the accuracy of the estimator at finite $N$
using bias defined by
\begin{align}
\mathrm{bias} = E_{\bm{Y},\bm{F}}[|\hat{\rho}(\bm{Y},\bm{F})-\rho|],
\end{align}
where $\hat{\rho}(\bm{Y},\bm{F})$ denotes the 
estimates under given $\bm{Y}$ and $\bm{F}$.
An accurate estimator results in a low bias value.

\item{(II)} Computational time

Although the mathematical forms of the estimator of prevalence
are similar,
the update rules in BP+EM algorithm and HBP algorithm differ from each other.
The BP+EM algorithm consists of a double loop, namely
the E-step for BP and the M-step for updating $\hat{\rho}$. 
In the HBP algorithm, the messages and the estimator are updated at the same time.
The difference between these update rules influences the 
computational time.

\end{description}

\begin{figure}
    \begin{minipage}{0.495\hsize}
    \centering
    \includegraphics[width=3in]{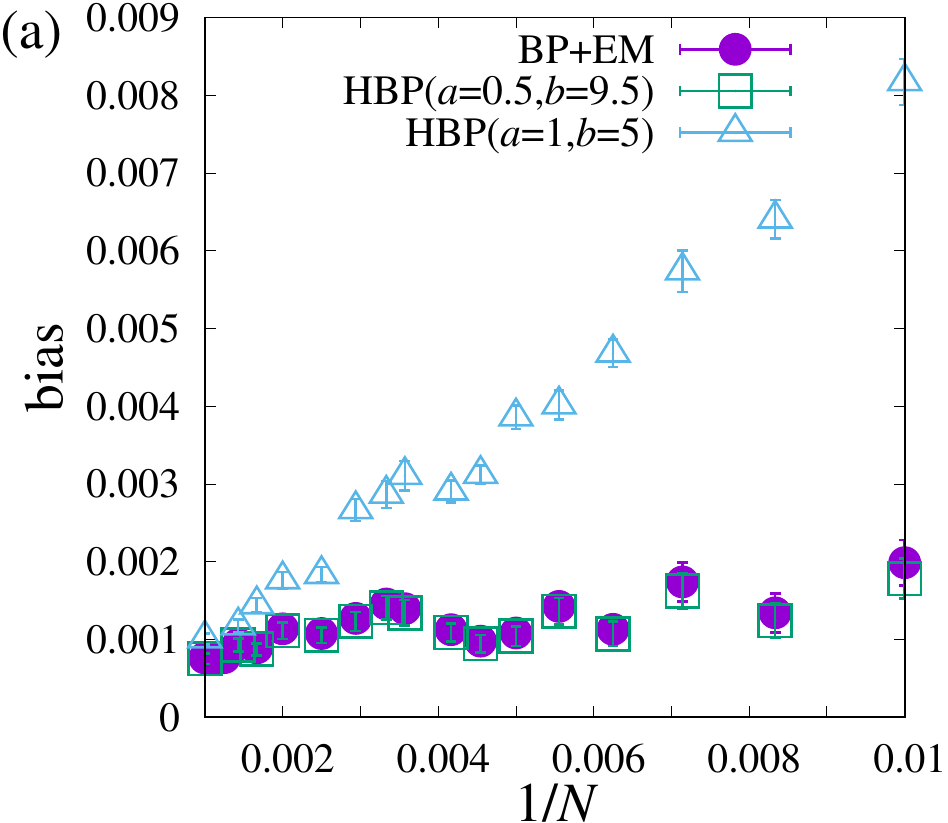}
    \end{minipage}
    \begin{minipage}{0.495\hsize}
    \centering
    \includegraphics[width=3in]{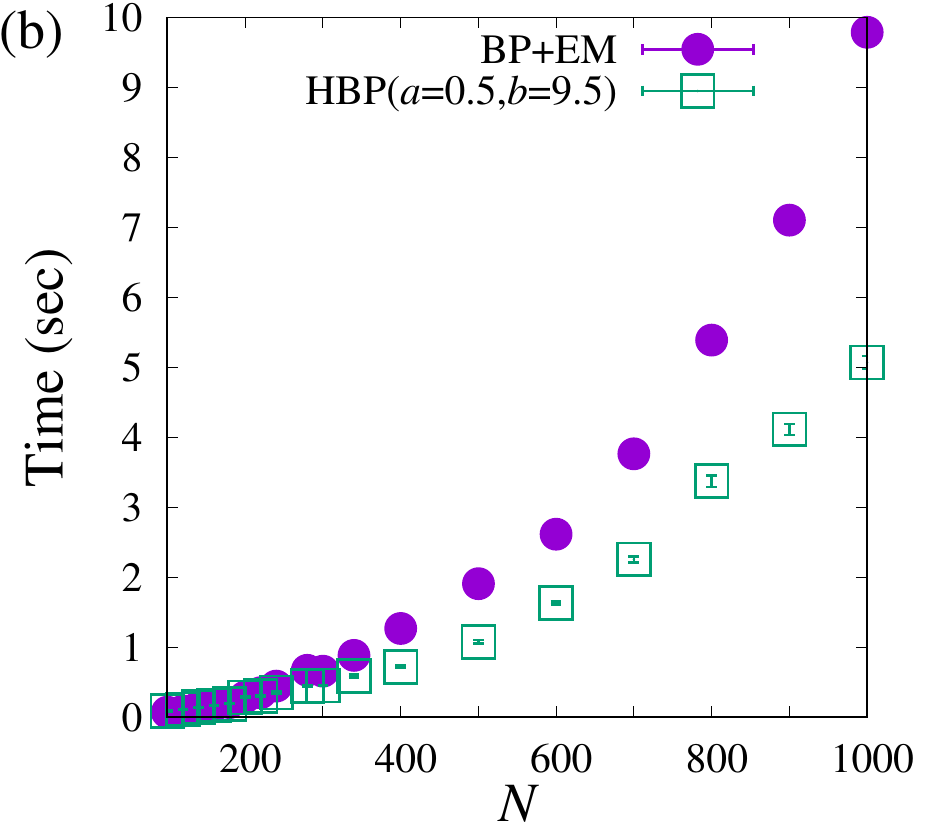}
    \end{minipage}
    \caption{(a) $1\slash N$-dependence of 
    bias and (b) $N$-dependence of computational time in seconds
    at $\alpha=0.5$, $\rho=0.05$, $p_{\mathrm{TP}}=0.99$, and $p_{\mathrm{FP}}=0.01$.
    }
    \label{fig:HBP_vs_EM}
\end{figure}

\begin{figure}
    \begin{minipage}{0.495\hsize}
    \centering
    \includegraphics[width=3in]{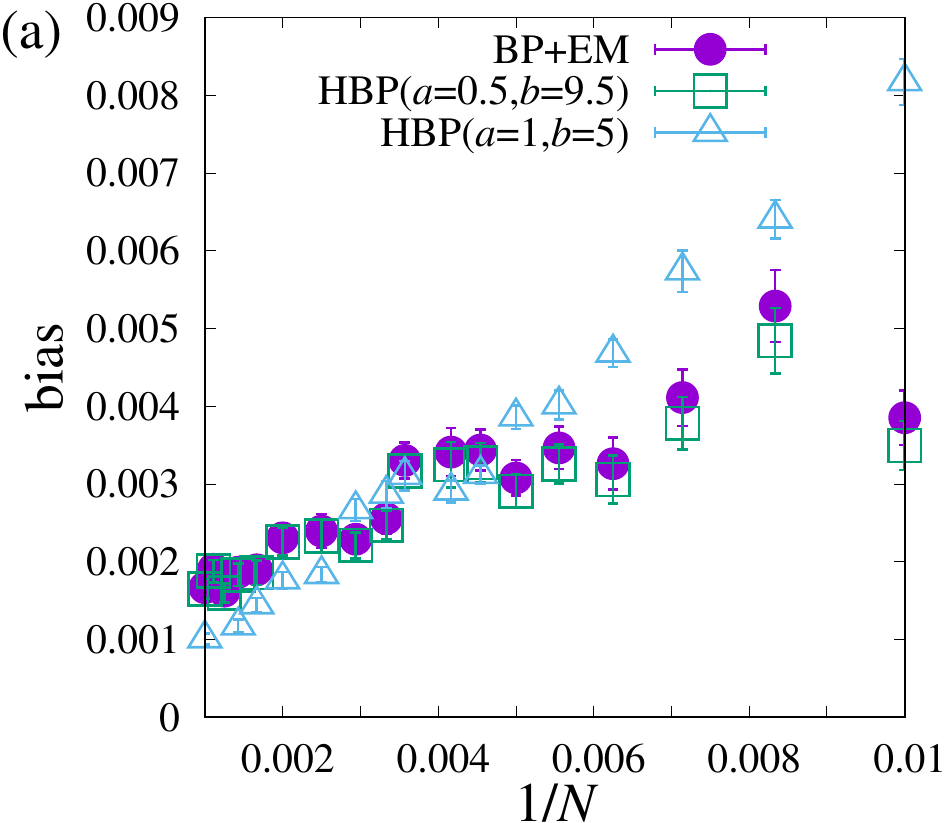}
    \end{minipage}
    \begin{minipage}{0.495\hsize}
    \centering
    \includegraphics[width=3in]{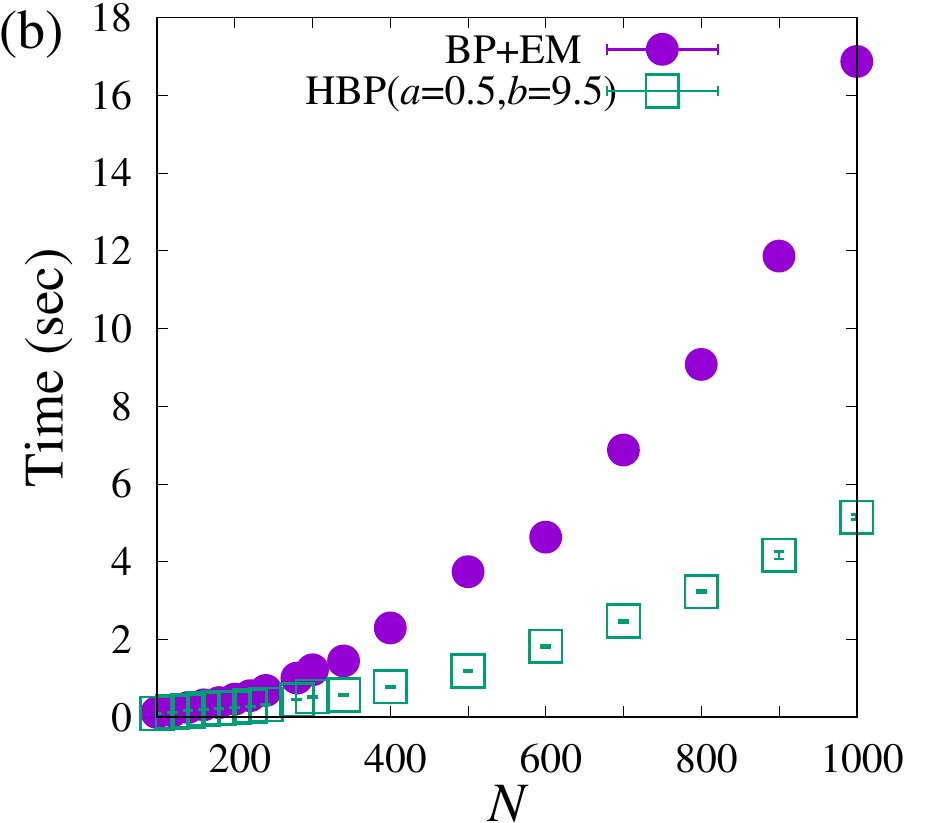}
    \end{minipage}
    \caption{(a) $1\slash N$-dependence of 
    bias and (b) $N$-dependence of computational time in seconds
    at $\alpha=0.5$, $\rho=0.05$, $p_{\mathrm{TP}}=0.95$, and $p_{\mathrm{FP}}=0.05$.
    }
    \label{fig:HBP_vs_EM2}
\end{figure}

\Fref{fig:HBP_vs_EM} (a) and \Fref{fig:HBP_vs_EM2} (a) show the
$N$-dependence of the bias 
for the BP+EM and HBP algorithms at $\alpha=0.5,~\rho=0.05,~p_{\mathrm{TP}}=0.99,
~p_{\mathrm{FP}}=0.01$ (\Fref{fig:HBP_vs_EM}),
and $\alpha=0.5,~\rho=0.05,~p_{\mathrm{TP}}=0.95,~p_{\mathrm{FP}}=0.05$ (\Fref{fig:HBP_vs_EM2}).
The same 100-realization of $\bm{X}^0$, $\bm{F}$, and $\bm{Y}$
were used for comparing these methods.
The mean of the hyperprior at $a=0.5,~b=0.95$ matches the true value of $\rho$.
BP+EM and HBP at $a=0.5,~b=0.95$ show almost the same dependency on $N$ in bias.
When $a$ and $b$ are not chosen to match the mean of the hyperprior,
bias becomes large in finite $N$, but the difference in bias
vanishes as $N\to\infty$.

\Fref{fig:HBP_vs_EM} (b) and \Fref{fig:HBP_vs_EM2} (b)
show the $N$-dependence of the computation time,
where we fixed our experimental environment to use a single 
3.5 GHz Intel Core i7 CPU.
The computational time of the HBP algorithm is less than that of the BP+EM algorithm,
and this priority stands out for the high-noise case,
which is evident from the comparison between 
\Fref{fig:HBP_vs_EM} (b) and \Fref{fig:HBP_vs_EM2} (b).

From these results, we consider that 
the choice of using the BP+EM or HBP algorithm depends on the purpose.
When precise estimation of prevalence is required
and there is no conception of the appropriate hyperprior in small system size,
the BP+EM algorithm should be used.
For quick identification of the infected patients, 
in particular for a large system size, the
HBP algorithm is well suited to the demand.

\section{Summary and discussion}
\label{sec:summary}

In this study, we investigated the group testing problem
where the test possesses finite false probabilities.
We introduced the BP algorithm to evaluate the infected patients
under the Bayesian inference settings.
The performance of the BP algorithm,
in particular for the TP rate, was improved by 
considering the credible interval of the point estimate
assigned to each patient.
Our approach used the bootstrap distribution 
to estimate the interval.
The unknown parameters in the model, particularly
prevalence, can be estimated 
using the EM method and hierarchical Bayes modeling.
We compared these methods and 
formulated a guide for practical usage.

We concentrated on
the pooling matrix randomly constructed under the column-wise
and row-wise constraints specified by $N_G$ and $N_O$.
The adaptive procedure of group testing was also examined,
where the pooling for the next stage was sequentially designed by
considering the output of the test in the previous stage \cite{Sobel1,Sobel2,AdaptiveGT}.
An extension of our BP and HBP algorithms to the adaptive setting is a 
promising way to explore more efficient pooling and test scheduling.

The MATLAB code used in this study is distributed on GitHub
\url{https://github.com/AyakaSakata/GroupTesting}.

\begin{acknowledgments}

This work was motivated by non-face-to-face discussions with Yukito Iba
during our remote work owing to COVID-19.
First, I thank him for our fruitful
discussions and his visionary comments.
Further, I thank Koji Hukushima, Yoshiyuki Ninomiya, Tomoyuki Obuchi, and Satoshi Takabe
for helpful comments and discussions.
This research was partially supported by a Grant-in-Aid for Scientific Research 19K20363 from the Japanese Society for the Promotion of Science (JSPS) and JST PRESTO Grant Number JPMJPR19M2, Japan.

\end{acknowledgments}

\appendix

\section{Comparison between exact calculation and approximation by BP algorithm}
\label{sec:exact}

\begin{figure}
    \begin{minipage}{0.495\hsize}
    \centering
    \includegraphics[width=3in]{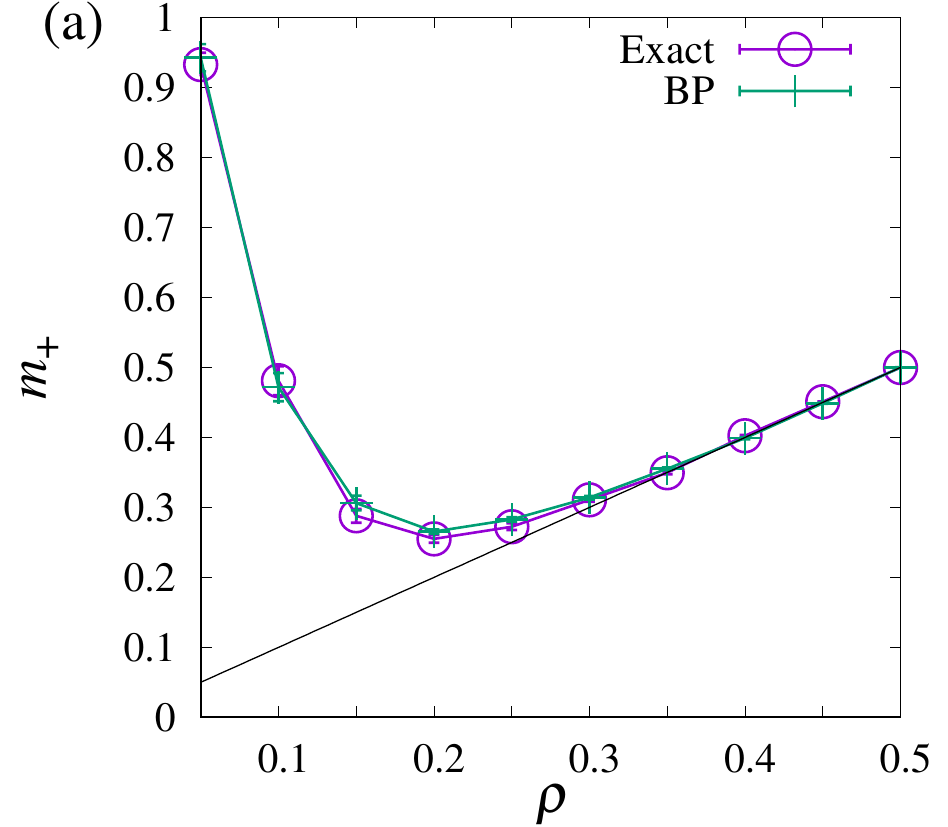}
    \end{minipage}
    \begin{minipage}{0.495\hsize}
    \centering
    \includegraphics[width=3in]{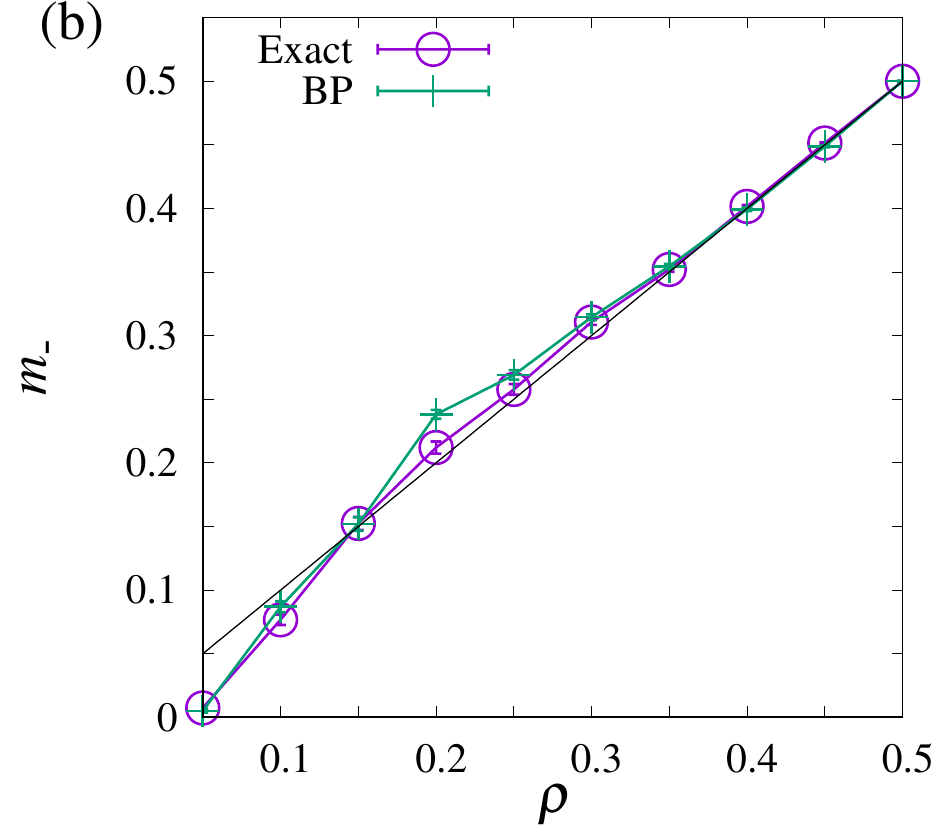}
    \end{minipage}
    \caption{$\rho$-dependence of (a) $m_+$ and (b) $m_-$ at $N=20, M=10$ and 
    $N_G=10$ for exact calculation (Exact) and approximation by BP algorithm (BP). 
    Error rates are fixed at $p_{\mathrm{TP}}=0.95$ and $p_{\mathrm{FP}}=0.02$.
    The gradients of solid lines are 1.}
    \label{fig:exact_vs_BP}
\end{figure}

To check the accuracy of the approximation by the BP algorithm, 
we perform exact calculation of the marginalized posterior distribution at $N=20$,
and compare it with the BP algorithm.
\Fref{fig:exact_vs_BP} shows (a) $m_+$ and (b) $m_-$ at 
$N=20$, $M=10$, $N_G=10$, $p_{\mathrm{TP}}=0.95$, and $p_{\mathrm{FP}}=0.02$,
calculated by sampling all possible configurations in $\bm{X}\in\{0,1\}^N$ (Exact)
and the BP algorithm (BP).
The gradients of the solid lines are 1.
The behaviors $m_+$ and $m_-$ derived by these two methods 
are similar to each other, and in particular, 
the tendency $m_+\to\rho$ and $m_-\to\rho$ as $\rho$ increases
is observed as discussed in Section\ref{sec:BP}.

\section{Derivation of maximum-(approximated) likelihood estimator}
\label{sec:EM_estimator}

The derivative of ${\cal S}$ with respect to $\rho$,
$p_{\mathrm{TP}}$, and $p_{\mathrm{FP}}$ is given by
\begin{align}
\nonumber
    \frac{\partial}{\partial\rho}{\cal S}&=\sum_i\frac{\partial}{\partial\rho}\ln{\cal Z}_i\\
    \nonumber
    &=\sum_i
    \frac{\prod_{\mu\in{\cal G}(i)}\tilde{\theta}_{\mu\to i} -\prod_{\mu\in{\cal G}(i)}(1-\tilde{\theta}_{\mu\to i})}{{\cal Z}_i}\\
    &=\frac{\sum_i\theta_i}{\rho}-\frac{\sum_i(1-\theta_i)}{1-\rho}\label{eq:dSdrho}\\
    \nonumber
    \frac{\partial}{\partial p_{\mathrm{TP}}}{\cal S}&=\sum_\mu\frac{\partial}{\partial p_{\mathrm{TP}}}\ln{\cal Z}_\mu\\
    \nonumber
    &=\sum_\mu\frac{Y_\mu(1-\tilde{q}_\mu)-(1-Y_\mu)(1-\tilde{q}_\mu)}{{\cal Z}_\mu}\\
    &=\frac{\sum_\mu\langle\mathbb{I}(Y_\mu=1,T_\mu(\bm{X})=1)\rangle_\mu}{p_{\mathrm{TP}}}-\frac{\sum_\mu\langle\mathbb{I}(Y_\mu=0,T_\mu(\bm{X})=1)\rangle_\mu}{1-p_{\mathrm{TP}}}\\
    \nonumber
     \frac{\partial}{\partial p_{\mathrm{FP}}}{\cal S}&=\sum_\mu\frac{\partial}{\partial p_{\mathrm{FP}}}\ln{\cal Z}_\mu\\
    \nonumber
    &=\sum_\mu\frac{Y_\mu\tilde{q}_\mu-(1-Y_\mu)\tilde{q}_\mu}{{\cal Z}_\mu}\\
    &=\frac{\sum_\mu\langle\mathbb{I}(Y_\mu=1,T_\mu(\bm{X})=0)\rangle_\mu}{p_{\mathrm{FP}}}-\frac{\langle\mathbb{I}(Y_\mu=0,T_\mu(\bm{X})=0)\rangle_\mu}{1-p_{\mathrm{FP}}},\label{eq:dSdw}
\end{align}
respectively.
Solving eqs. (\ref{eq:dSdrho})--(\ref{eq:dSdw}) under the condition that they are zero,
we obtain the extremization conditions in eqs.(\ref{eq:rhoh})--(\ref{eq:w_hat}).

\section{Estimated value of the prevalence in hierarchical Bayes model
at $N\to\infty$}
\label{sec:saddle}

Substituting \eref{eq:s} into \eref{eq:xih},
we obtain the following expression.
\begin{align}
\tilde{\rho}_i=\frac{\int d\rho \rho\phi(\rho)\exp \left[N\left\{\frac{1}{N}\sum_{j\neq i}\log\{\rho\pi_i+(1-\rho)(1-\pi_i)\}\right\}\right]}{\int d\rho\phi(\rho)\exp \left[N\left\{\frac{1}{N}\sum_{j\neq i}\log\{\rho\pi_i+(1-\rho)(1-\pi_i)\}\right\}\right]}.
\end{align}
Applying the saddle point method,
we obtain
\begin{align}
\tilde{\rho}_i=\rho_i^*,
\end{align}
where $\rho_i^*$ satisfies 
\begin{align}
\frac{1}{N}\sum_{j\neq i}\frac{\pi_i-(1-\pi_i)}{\rho_i^*\pi_i+(1-\rho_i^*)(1-\pi_i)}=0.
\label{eq:saddle_rho}
\end{align}
From eqs. (\ref{eq:xi}) and (\ref{eq:theta_tilde}),
\eref{eq:saddle_rho} is transformed as
\begin{align}
\frac{1}{N}\sum_{j\neq i}\frac{\tilde{\theta}_j(\rho_i^*)}{\rho_i^*} = \frac{1}{N}\sum_{j\neq i}\frac{1-\tilde{\theta}_j(\rho_i^*)}{1-\rho_i^*},
\label{eq:saddle_rho2}
\end{align}
and we obtain \eref{eq:rhoh_HBP}
by transforming \eref{eq:saddle_rho2} with respect to $\rho_i^*$.

\providecommand{\noopsort}[1]{}\providecommand{\singleletter}[1]{#1}%

\end{document}